\documentclass[11pt]{article}

\usepackage[preprint]{acl}

\usepackage{times}
\usepackage{latexsym}

\usepackage[T1]{fontenc}

\usepackage[utf8]{inputenc}

\usepackage{microtype}

\usepackage{inconsolata}

\usepackage{graphicx}
\usepackage{cuted}

\usepackage{multirow}
\usepackage{xspace}

\usepackage{amsmath}
\usepackage{multirow}
\usepackage{bbm}
\usepackage{amssymb}
\usepackage{booktabs}
\usepackage{lscape}
\usepackage{pifont}
\usepackage{algorithm}
\usepackage{caption}
\usepackage{subcaption}
\usepackage{CJKutf8}
\usepackage{makecell}
\usepackage{xcolor}
\usepackage{listings}
\usepackage{xcolor}
\usepackage[most]{tcolorbox}
\usepackage{multirow}

\usepackage{amsmath}
\usepackage{multirow}
\usepackage{bbm}
\usepackage{graphicx}
\usepackage{amssymb}
\usepackage{booktabs}
\usepackage{lscape}
\usepackage{pifont}
\usepackage{algorithm}
\usepackage{caption}
\usepackage{subcaption}
\usepackage{CJKutf8}
\usepackage{makecell}
\usepackage{xcolor}
\usepackage{svg}
\usepackage{listings}
\usepackage{xcolor}
\usepackage{kotex}
\usepackage{longtable}
\usepackage{fvextra}

\newcommand{\RQone}{\texttt{RQ1}\xspace}
\newcommand{\RQtwo}{\texttt{RQ2}\xspace}
\newcommand{\RQthree}{\texttt{RQ3}\xspace}

\newcommand\datasetname{\textcolor{black}{\textsc{CodeMixQA}}}

\newtcolorbox[auto counter, number within=section]{promptbox}[2][]{
    enhanced,
    breakable,
    colback=gray!5,
    colframe=black!80,
    boxrule=0.5pt,
    arc=2pt,
    left=8pt,right=8pt,top=8pt,bottom=8pt,
    title={\thetcbcounter:~#2},
    fonttitle=\sffamily\bfseries,
    coltitle=black,
    colbacktitle=gray!15,
    #1
}

\title{Can Large Language Models Understand, Reason About, and Generate Code-Switched Text?}

\author{Genta Indra Winata$^1$, David Anugraha$^2$, Patrick Amadeus Irawan$^3$, Anirban Das$^1$,\\
\textbf{Haneul Yoo$^4$, Paresh Dashore$^1$, Shreyas Kulkarni$^1$, Ruochen Zhang$^5$, Haruki Sakajo$^6$,}\\
\textbf{Frederikus Hudi$^6$, Anaelia Ovalle$^7$, Syrielle Montariol$^8$, Felix Gaschi$^9$,}\\
\textbf{Michael Anugraha$^{7}$, Rutuj Ravindra Puranik$^{10}$, Zawad Hayat Ahmed$^{7}$,}\\
\textbf{Adril Putra Merin$^{11}$, Emmanuele Chersoni$^{12}$} \\
$^1$Capital One$\quad$$^2$Stanford University$\quad$$^3$MBZUAI$\quad$$^4$KAIST$\quad$$^5$Brown University \\
$^6$NAIST$\quad$$^7$Independent$\quad$$^8$UC Berkeley$\quad$$^9$SAS Posos\\
$^{10}$Oracle$\quad$$^{11}$ITB$\quad$$^{12}$The Hong Kong Polytechnic University \\
\small\texttt{genta.winata@capitalone.com, david.anugraha@stanford.edu, patrick.irawan@mbzuai.ac.ae}}

\begin{document}
\maketitle
\begin{abstract}
Code-switching is a pervasive phenomenon in multilingual communication, yet the robustness of large language models (LLMs) in mixed-language settings remains insufficiently understood. In this work, we present a comprehensive evaluation of LLM capabilities in understanding, reasoning over, and generating code-switched text. We introduce $\datasetname$, a novel benchmark with high-quality human annotations, comprising 16 diverse parallel code-switched language-pair variants that span multiple geographic regions and code-switching patterns, and include both original scripts and their transliterated forms. Using this benchmark, we analyze the reasoning behavior of LLMs on code-switched question-answering tasks, shedding light on how models process and reason over mixed-language inputs. We further conduct a systematic evaluation of LLM-generated synthetic code-switched text, focusing on both naturalness and semantic fidelity, and uncover key limitations in current generation capabilities. Our findings reveal persistent challenges in both reasoning and generation under code-switching conditions and provide actionable insights for building more robust multilingual LLMs. We release the dataset and code as open source.\footnote{The dataset can be found at~\url{https://huggingface.co/datasets/gentaiscool/codemixqa} and the code can be found at~\url{https://github.com/gentaiscool/codemixqa}.}
\end{abstract}

\section{Introduction}

\begin{figure}[!th]
    \centering
    \includegraphics[width=0.49\textwidth]{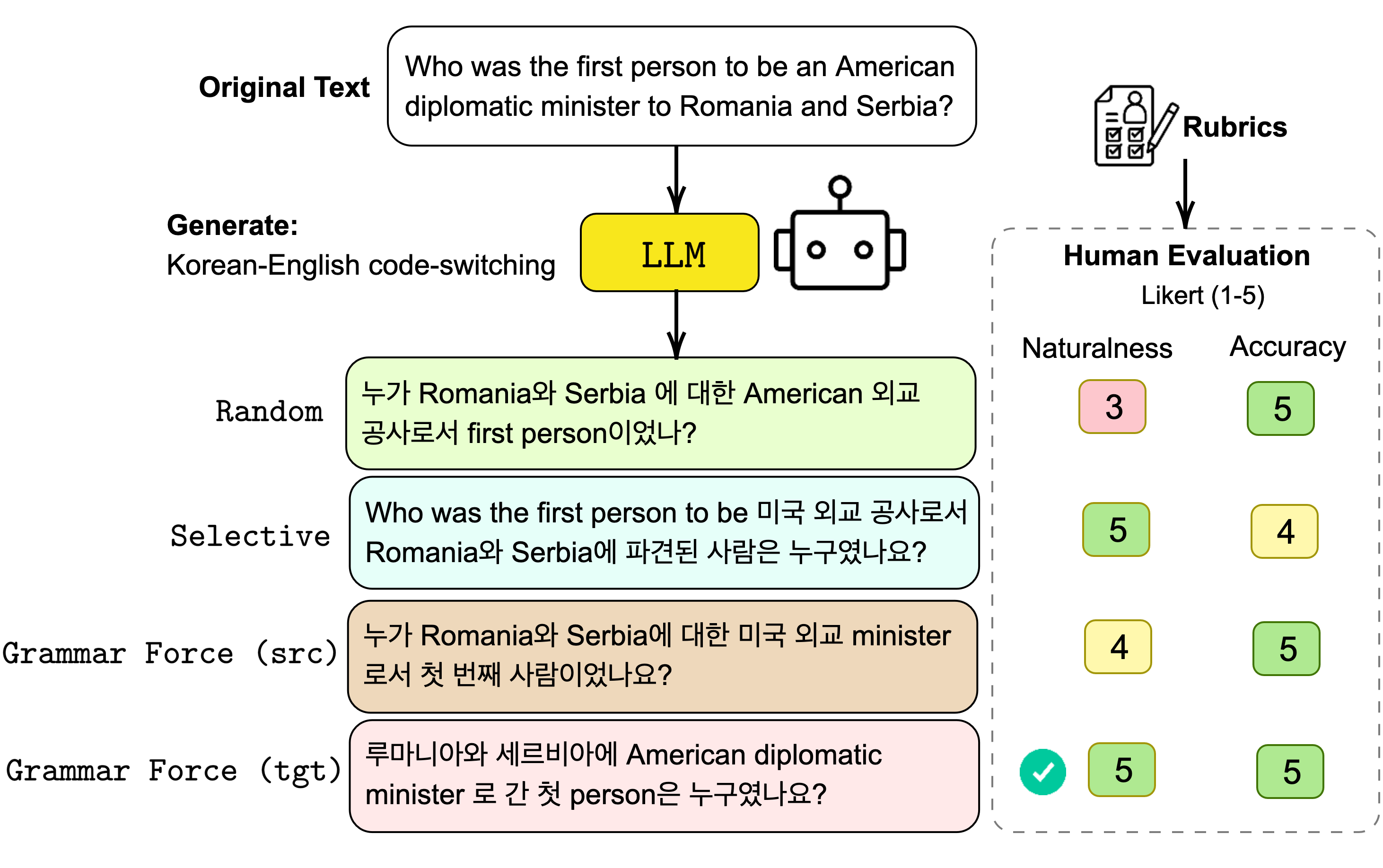}
    \caption{Example of code-switching generation for Korean-English using different synthetic data methods. It also presents the human evaluation results leveraging the rubrics, where annotators showed a preference for \textsc{Grammar Force (Tgt)}.}
    \label{fig:illustration.}
\end{figure}

Robustness in large language models (LLMs) is a critical area of study, particularly in evaluating whether these models can perform reliably in multilingual settings that reflect real-world language use~\cite{myung2024blend,yong2025crosslingual}. In many multilingual communities, communication frequently involves code-switching, a highly prevalent linguistic phenomenon~\cite{myers2001matrix,winata2021multilingual,winata2023decades}. Consequently, developing models that can accurately identify, interpret, and respond to mixed-language inputs is essential for achieving robust language understanding. Although recent LLMs have demonstrated some capacity to handle code-switching, earlier and legacy models were often only partially successful, frequently failing to fully comprehend mixed-language inputs or to generate appropriate responses~\cite{zhang2023multilingual}.

In this study, we would like to answer the following research questions:
\begin{enumerate}
    \item[\RQone.] \textbf{Can LLMs understand code-switching?} We evaluate the robustness of LLMs in understanding and responding to code-switched inputs across downstream NLP tasks.
    \item[\RQtwo.] \textbf{Can LLMs reason over code-switched inputs?} We investigate how LLMs perform reasoning when processing code-switched text and examine whether introducing mixed languages leads to misalignment or errors in their reasoning process.
    \item[\RQthree.] \textbf{Can LLMs generate code-switched text?} We examine the ability of LLMs to generate code-switched texts and assess the naturalness and semantic quality of the generated outputs.
\end{enumerate}
        
Our contribution can be summarized as three-fold as follows:
\begin{itemize}
    \item To initiate the study, we introduce $\datasetname$, a novel code-switching dataset with high-quality human annotations, comprising 16 diverse parallel code-switched language-pair variations in original scripts and their transliterations, designed to evaluate LLM performance across different code-switching patterns and text generation strategies.
    \item We analyze the reasoning behavior of LLMs on code-switching question-answering tasks, providing insights into how models process and reason over mixed-language inputs. Our result reveal an average performance degradation of approximately 11\% under code-switching conditions, with higher code-mixing density correlating with improved task robustness.
    \item We conduct a systematic analysis of the naturalness and semantic accuracy of synthetic code-switched text generated by LLMs, uncovering key limitations in current code-switching generation capabilities. Our findings show that outputs preserving the non-English language as the matrix language are rated as significantly more natural by native speakers, whereas English-dominant grammatical structures achieve higher task performance but lower naturalness, highlighting a fundamental tradeoff in code-switching generation. Interestingly, LLMs perform better on perturbed text when English serves as the matrix language, suggesting a stronger proficiency in handling English-dominant inputs.
\end{itemize}

\begin{table*}[!th]
\centering
\resizebox{\textwidth}{!}{
    \begin{tabular}{l|cccccccc}
    \toprule
    \textbf{Dataset} & \textbf{\# Lang. Pair} & \textbf{\# Size} & \textbf{Synthesis Method} & \textbf{Source Data} & \textbf{LLM Eval.} & \textbf{Human Eval.} & \textbf{Open-Source} & \textbf{License} \\ 
    & \textbf{Variations} \\ \midrule
    \citet{pratapa2018language} & 1 & 31M & Sentence Parsing with Linguistic Theory & Parallel Data & $\times$ & $\checkmark$ (No details) & $\times$ & N/A \\ 
    PointerGen~\cite{winata2019code} & 1 & 270k & Finetune and Inference & Parallel Data & $\times$ & $\times$ & $\times$ & N/A \\
    GLUECoS~\cite{khanuja2020gluecos} & 2 & 10M$^\dagger$ & \citet{pratapa2018language} & Parallel Data & $\times$ & $\times$ & $\times$ & MIT \\
    CSPref~\cite{kuwanto2024linguistics} & 3 & 69k & LLM with Linguistic Theories & Parallel Data & $\times$ & $\checkmark$  & Unknown & N/A\\
    \citet{yong2023prompting} & 7 & Unknown & LLM Prompting & Any & $\times$ & $\checkmark$  & $\times$ & N/A \\
    AfroCS-xs~\cite{olaleye2025afrocs} & 4 & 3.6k & LLM with Few-shot ICL & Any & $\times$ & $\checkmark$ (partially) & $\times$ & N/A \\
    CodeMixBench~\cite{yang2025codemixbench} & 12$^\ddagger$ & 64k & Random Word Replacement & Parallel Data & $\checkmark$ & $\times$  & $\checkmark$ & Apache 2.0\\ 
    SwitchLingua~\cite{xie2025switchlingua} & 12 & 80 hrs (speech) & LLM with Linguistic Theories & Any & $\checkmark$ & $\checkmark$ & $\checkmark$ & CC-BY-NC 4.0 \\ \midrule
    $\datasetname$ & 16 & 64k & LLM with Linguistic Theories & Any & $\checkmark$ & $\checkmark$ & $\checkmark$ &  CC-BY-SA 4.0 \\
    \bottomrule
    \end{tabular}
}
\caption{Comparison of Synthetic Code-Switching Datasets. $\datasetname$ covers a more diverse set of languages and is larger in scale than existing benchmarks. It supports comprehensive LLM evaluation through both downstream task performance and native-speaker human evaluation, and will be released under a permissive license. In addition, the dataset includes transliterated variants as distinct language-pair configurations.\\
\emph{Note:} 
$^\dagger$The 10M-sentence synthetic component (gCM) is not publicly available. 
$^\ddagger$The newly generated synthetic portion contains 12 languages; the total of 18 languages is achieved by incorporating external datasets (GLUECoS and LINCE), which are excluded from our analysis.}
\label{tab:compare-benchmark}
\end{table*}

\section{Background and Related Work}



Early work on code-switching generation primarily relies on parallel data and statistical models, leveraging the Functional Head Constraint~\cite{li2012code,li2014language}. This is followed by rule-based grammatical approaches that draw on linguistic theories to produce linguistically grounded code-switched text~\cite{bhat2016grammatical,pratapa2018language}. \citet{winata2019code} subsequently introduce the first neural-based approach for code-switching generation, training models to learn code-switching behavior using a copy mechanism within a neural machine translation (NMT) framework; this approach is later extended by \citet{gupta2020semi} to additional language pairs. More recently, Generative Adversarial Network (GAN)-based methods~\cite{chang2019code,chai2021counter} and Variational Autoencoders (VAEs)~\cite{samanta2019deep} have been explored for code-switched data synthesis. Another line of work focuses on adapting pre-trained large language models (LLMs) for code-switching generation~\cite{hsu2023code}.

\citet{yang2025codemixbench} construct a code-switching dataset using a purely model-based synthesis approach; however, their method relies primarily on word-level replacement and does not condition generation on grammatical structure. Consequently, it offers limited insight into generating natural-sounding code-switched text, particularly in the absence of human evaluation. In contrast, SwitchLingua introduces an LLM-based approach for generating code-switched data to support multilingual and multi-ethnic dataset construction~\cite{xie2025switchlingua}, but it focuses exclusively on speech data for training automatic speech recognition models. As a result, its scope is restricted to speech recognition applications and does not address code-switching generation or evaluation in broader NLP contexts. Additionally, the human annotation rubric is relatively non-standard, and it is unclear how linguistic theory is incorporated, as this is not explicitly discussed.

A key limitation of prior work is that it focuses primarily on generating synthetic code-switched data, without investigating how such data affects model performance or the robustness of LLMs on downstream tasks. Additionally, there is still a lack of understanding of how LLMs reason over code-switched inputs and how well they handle mixed-language conditions. In this work, we address this gap by introducing a dataset that remains challenging for LLMs to evaluate even in English at the time of its creation, providing a more rigorous benchmark for assessing model comprehension and reasoning in multilingual contexts.

\section{CodeMixQA}
To initiate our study, we introduce a new code-switching dataset. This section describes the dataset construction considerations and process, including the steps involved and the different methods used to generate the data.

\subsection{Design Consideration}
\paragraph{Transliteration and Evaluation.} Designing a code-switching dataset requires careful consideration of linguistic and cultural factors. In some multilingual communities, language mixing follows specific writing conventions. For instance, in Indian languages such as Bengali, Hindi, Marathi, and Urdu, code-switching is often expressed in romanized form rather than in the native script when mixed with English, primarily for ease of typing. In contrast, in languages such as Chinese, Japanese, and Korean, code-switching typically retains the original script, as it is more natural in those contexts. Accordingly, we construct both transliterated and native-script versions of the dataset to examine model preferences for Indian languages. Additionally, we conduct evaluations with native speakers to assess the naturalness of the generated code-switched text.

\paragraph{Open-Source.} We release the dataset under the CC-BY-SA 4.0 license, which permits both reuse and redistribution while requiring appropriate attribution. This license ensures that the dataset can be safely used by the research community for future studies, including adaptation, extension, and benchmarking of new models. By adopting a permissive yet responsible licensing framework, we aim to facilitate reproducibility, encourage collaboration, and support the ongoing development of code-switching research in multilingual NLP.

\begin{table*}[!th]
\centering
\resizebox{\textwidth}{!}{
\begin{tabular}{lcccccccccc}
\toprule
\textbf{ISO Code} & \textbf{Languages} & \textbf{Transliterate} & \multicolumn{2}{c}{\textbf{Random}} & \multicolumn{2}{c}{\textbf{Selective}} & \multicolumn{2}{c}{\textbf{Grammar Force (Src)}} & \multicolumn{2}{c}{\textbf{Grammar Force (Tgt)}} \\ 
& & & CMI & SPF & CMI & SPF & CMI & SPF & CMI & SPF\\ \midrule
\texttt{bbc-eng} & Toba Batak - English & $\times$ & 0.74 & 0.41 & 0.76 & 0.42 & 0.76 & 0.41 & 0.77 & 0.43 \\
\texttt{ben-eng} & Bengali - English & $\checkmark$ & 0.67 & 0.37 & 0.64 & 0.34 & 0.60 & 0.32 & 0.60 & 0.32 \\
\texttt{esp-eng} & Spanish - English & $\times$ & 0.64 & 0.35 & 0.61 & 0.33 & 0.73 & 0.42 & 0.48 & 0.26 \\
\texttt{fra-eng} & French - English & $\times$ & 0.64 & 0.34 & 0.62 & 0.33 & 0.74 & 0.41 & 0.61 & 0.33 \\
\texttt{hin-eng} & Hindi - English & $\checkmark$ & 0.45 & 0.25 & 0.55 & 0.30 & 0.35 & 0.19 & 0.54 & 0.29 \\
\texttt{ind-eng} & Indonesian - English & $\times$ & 0.69 & 0.37 & 0.64 & 0.34 & 0.72 & 0.40 & 0.58 & 0.32 \\
\texttt{ita-eng} & Italian - English & $\times$ & 0.72 & 0.40 & 0.70 & 0.39 & 0.76 & 0.44 & 0.64 & 0.36 \\
\texttt{jpn-eng} & Japanese - English & $\times$ & 0.46 & 0.23 & 0.35 & 0.15 & 0.63 & 0.30 & 0.20 & 0.09 \\
\texttt{kor-eng} & Korean - English & $\times$ & 0.35 & 0.17 & 0.30 & 0.15 & 0.60 & 0.28 & 0.25 & 0.13 \\
\texttt{mar-eng} & Marathi - English & $\checkmark$ & 0.78 & 0.44 & 0.78 & 0.42 & 0.67 & 0.37 & 0.72 & 0.38 \\
\texttt{urd-eng} & Urdu - English & $\checkmark$ & 0.74 & 0.40 & 0.69 & 0.35 & 0.61 & 0.32 & 0.68 & 0.35 \\
\texttt{zho-eng} & Chinese - English & $\times$ & 0.58 & 0.27 & 0.53 & 0.24 & 0.60 & 0.25 & 0.32 & 0.15 \\
\bottomrule
\end{tabular}
}
\caption{Dataset languages and code-switching statistics for CMI and SPF across different generation strategies.}
\label{tab:languages}
\end{table*}

\subsection{Data Creation}\label{subsection:data_creation}

Creating code-switching data using LLMs is non-trivial, as the resulting text must not only exhibit language switching but also remain natural and semantically accurate. We use SimpleQA Verified~\cite{haas2025simpleqa} as our source dataset. We select the SimpleQA Verified, as it is a challenging evaluation set that has not been saturated yet by current models and has desirable properties such as verifiable answers (through source reconciliation), de-duplicated data points, topic balancing, and that it is markedly different from most standard tasks that are prevalent in code switching studies such as language identification, NER, and machine translation~\cite{winata2023decades}.
In this dataset, we employ multiple data generation strategies, including random switching, selective switching, and grammar-constrained approaches. We extend the dataset to 16 code-switched language variations that are diverse across geographic regions (Table~\ref{tab:languages}). For simplicity, English is treated as the embedded language (L2), while each non-English language serves as the matrix language (L1). Code-switched text is generated using multiple LLM prompting strategies, including \textsc{Random Switching}, which permits a predefined maximum number of language switches; \textsc{Selective Switching}, where LLMs are instructed to produce natural-sounding code-switched text without additional linguistic constraints; and \textsc{Grammar Forcing}, a linguistically inspired approach that guides LLMs to preserve the grammatical structures of both the matrix and embedded languages, drawing on the Matrix Language Frame and Equivalence Constraint theories~\cite{myers2001matrix}. In this process, we alternate the source (src) and target (tgt) languages as the matrix and embedded languages. Compared to prior approaches~\cite{winata2019code,kuwanto2024linguistics}, our method provides a more practical and scalable solution by explicitly encoding linguistic constraints as prompt-level instructions for LLMs, rather than relying on alignment learned from small datasets or translation-based methods, which are often brittle and fail to generalize to unseen code-switching patterns.

\paragraph{Random Switching.} We generate code-switched text by instructing LLMs to randomly replace words or phrases in the original English text. No additional constraints are imposed, allowing the models to switch languages freely. We consider three settings that cap the maximum proportion of switched content at 50\%.

\paragraph{Selective Switching.} We generate code-switched text by conditioning the model to produce natural code-switching. Natural switching is defined as preserving the original semantics of the text while maintaining linguistic naturalness and fluency.

\paragraph{Grammar Forcing.} We generate code-switched text by following the \textsc{Selective Switching} criterion to ensure naturalness, with additional constraints that enforce the preservation of either English or non-English grammatical structure. This notion of grammar forcing is inspired by the Equivalence Constraint theory, which emphasizes preserving grammatical structure during code-switching~\cite{myers2001matrix}.

\noindent For further details on the prompts, please refer to Appendix~\ref{sec:prompts}.


\section{Experiments}

\subsection{Tasks and Setup}
We evaluate three core capabilities of LLMs in code-switching settings: understanding, reasoning, and generation, using the following strategies.

\paragraph{Understanding.}
We assess LLM understanding by evaluating their ability to answer perturbed (code-switched) questions, comparing performance on these questions to their non-perturbed English counterparts. This evaluation measures the robustness of LLMs to mixed-language inputs under various code-switching conditions and examines their language-agnostic capabilities, that is, whether the models’ performance degrades due to language mixing. We report F1 score as the primary evaluation metric.

\paragraph{Reasoning.}
To assess reasoning ability, we analyze the models’ reasoning traces and examine how their behavior changes when code-switching perturbations are applied to the input. We categorize reasoning issues using an LLM as a judge, following the nine categories defined in~\citet{ovalle2025beg}. Additionally, we provide definitions to guide the LLM judge in evaluating misalignment across different languages.

\paragraph{Generation.}
We evaluate the quality of LLM generated text through human evaluation. We randomly sample 100 questions from SimpleQA and generate four perturbed versions for each language. For each language pair, we recruit native speakers as annotators to assess the generated code switched text along two dimensions: (1) naturalness and (2) semantic accuracy. All evaluations are conducted using a 5 point Likert scale, with detailed annotation guidelines provided in Appendix~\ref{sec:annotation-rubrics}. In addition to human evaluation, we compute two code switching metrics Code Mixing Index (CMI) and Switch Point Fraction (SPF) for the generated text, following \citet{winata2019code}.

\subsection{Generation Code-Switching Metrics}
To systematically evaluate the quality of code-switched text generated by LLMs, it is important to quantify both the extent of language mixing and the distribution of switch points within utterances. We adopt two widely used metrics, the Code-Mixing Index (CMI)~\cite{gamback2016comparing,winata2019code} and the Switch-Point Fraction (SPF)~\cite{pratapa2018language}, to capture these aspects, providing a comprehensive view of code-switching behavior in generated text. These metrics, combined with preprocessing steps for tokenization and language identification, allow us to rigorously assess both the structure and naturalness of mixed-language outputs.

\paragraph{Code-Mixing Index (CMI).}
The Code-Mixing Index quantifies the degree of code-mixing in an utterance by measuring the extent to which multiple languages are used. At the utterance level, CMI is computed by first identifying the dominant language (i.e., the most frequent language in the utterance) and then counting the tokens from all other languages. Following prior work, we compute corpus-level CMI by averaging utterance-level scores across all sentences. Formally, for an utterance $x$, CMI is defined as:
\begin{equation}
    CMI(x) = \frac{N(x) - \max_{\ell_i \in \ell} \left\{ t_{\ell_i}(x) \right\} + P(x)}{N(x)},
\end{equation}
where $N(x)$ is the total number of tokens in $x$, $t_{\ell_i}(x)$ is the number of tokens in language $\ell_i$, and $P(x)$ is the number of code-switch points. The corpus-level CMI is obtained by averaging $CMI(x)$ over all utterances.

\begin{table*}[!th]
\centering
\resizebox{\textwidth}{!}{
\begin{tabular}{lc|cccc|c}
\toprule
\textbf{Model} 
& \textbf{No Perturb.} & \textbf{Random} & \textbf{Selective} & \multicolumn{1}{c}{\textbf{Grammar Forcing (Src)}} & \textbf{Grammar Forcing (Tgt)} & \textbf{avg.} \\ \midrule
\textsc{Gemma3 12B} & 8.90 & 8.12 & \textbf{8.22} & \underline{8.13} & 7.91 & 8.10 \\
\textsc{Gemma3 27B} & 13.00 & \textbf{11.50} & \underline{11.47} & 11.37 & 11.18 & 11.38 \\
\textsc{Llama3.3 70B Instruct} & 19.50 & 16.99 & \textbf{17.81} & 16.35 & \underline{17.67} & 17.21 \\ \midrule
\textsc{Qwen3 4B} & 8.40 & \underline{8.24} & 8.21 & \textbf{8.89} & 8.16 & 8.38 \\
\textsc{Qwen3 30B A3B} & 23.70 & \underline{20.89} & 20.57 & \textbf{21.45} & 20.44 & 20.84 \\
\textsc{Olmo 3 7B} & 5.40 & \underline{5.29} & \textbf{5.30} & 4.89 & 5.15 & 5.16 \\
\textsc{Olmo 3 32B} & 7.60 & 7.44 & \underline{7.58} & 7.49 & \textbf{7.78} & 7.57 \\
\textsc{Olmo 3.1 32B} & 8.60 & 6.44 & \textbf{6.76} & \underline{6.58} & \underline{6.58} & 6.59 \\
\textsc{GPT-OSS 20B} & 22.30 & \underline{16.33} & 15.94 & \textbf{17.81} & 15.84 & 16.48 \\
\textsc{GPT-OSS 120B} & 28.30 & 21.59 & \underline{22.17} & \textbf{24.08} & 21.09 & \underline{22.23} \\
\textsc{GPT-4.1} & 40.10 & 36.79 & 36.35 & \textbf{37.32} & \underline{36.86} & \textbf{36.83} \\ \midrule
\textsc{avg.} & 16.89 &	14.51 &	\underline{14.58} & \textbf{14.94} &	14.42 \\
\bottomrule
\end{tabular}
}
\caption{LLM results in F1 score under different perturbation strategies. The \texttt{avg.} column reports the average performance across perturbation settings only, excluding the non-perturbed condition.}
\label{tab:results_perturbation}
\end{table*}

\paragraph{Switch-Point Fraction (SPF).}
SPF measures the density of language switches within a sentence by computing the ratio of switch points to the total number of word boundaries. A \emph{switch point} is defined as a boundary between two adjacent words where the languages differ. Formally, for an utterance $x$:
\begin{equation}
    SPF(x) = \frac{P(x)}{N(x)-1},
\end{equation}
where $P(x)$ and $N(x)$ are defined as in CMI.

To assess text quality, we identify code-switching points and determine the language of each word, which requires several preprocessing steps. For scripted text, we first apply tokenization, as generated text in some scripts may lack explicit word boundaries. We then apply a language identification classifier to assign a language label to each token. Further details are provided in Appendix~\ref{sec:preprocessing}.

\subsection{Data Statistics}
For each language, we generate four text variants per sample (one using \textsc{Random Switching}, one using \textsc{Selective Switching}, and two using \textsc{Grammar Forcing}), resulting in 4k samples per language. Across 16 languages, this yields a total of 64k samples.

Table~\ref{tab:languages} shows the CMI and SPF metrics for each code-switching language variation. The results show that East Asian and non-Latin-scripted languages, such as Japanese, Korean, Hindi, and Chinese, have a lower complexity baseline in both the random and selective modes. This indicates that the lower natural complexity of their unique grammars limits word-level transitions with English. Applying \textsc{Grammar Force (Src)} consistently produces the highest complexity by forcing secondary language tokens into the source language's structure, which maximizes both token density (via CMI) and the frequency of switches (via SPF). In contrast, \textsc{Grammar Force (Tgt)} results in the lowest complexity, suggesting that the target language grammar is more restrictive about where switches can occur, thereby creating a more limited mixing pattern.

\subsection{Models}
\paragraph{Understanding and Reasoning.} To evaluate LLM comprehension capabilities, we assess nine open-weight models: \textsc{Gemma3} (\textsc{12B} and \textsc{27B})~\cite{team2025gemma}, \textsc{Llama3.3 70B Instruct}~\cite{grattafiori2024llama}, \textsc{Qwen3} (\textsc{4B} and \textsc{30B A3B})~\cite{yang2025qwen3}, \textsc{Olmo} (\textsc{3.1 32B}, \textsc{3 7B}, and \textsc{3 32B})~\cite{olmo2025olmo}, and \textsc{GPT-OSS} (\textsc{20B} and \textsc{120B})~\cite{agarwal2025gpt}. Due to resource constraints, we include only one proprietary model, \textsc{GPT-4.1}. For reasoning evaluations, we focus exclusively on thinking-oriented model variants, excluding instruct models from the analysis. Additionally, we use \textsc{GPT-OSS 120B} to assess reasoning trace issues following perturbation.

\begin{table*}[!th]
\centering
\resizebox{\textwidth}{!}{
\begin{tabular}{lccccccccccccc}
\toprule
\textbf{Model} 
& \textbf{No Perturb.} & \textbf{bbc-eng} & \multicolumn{2}{c}{\textbf{ben-eng}} & \textbf{esp-eng} & \textbf{fra-eng} & \multicolumn{2}{c}{\textbf{hin-eng}} & \textbf{ind-eng} \\
& & & ori. & trans. & & & ori. & trans. &  \\ \midrule
\textsc{Gemma3 12B} & 8.90 & 11.10 & 7.40 & 9.70 & 7.62 & 7.52 & 8.30 & 8.08 & 8.30  \\
\textsc{Gemma3 27B} & 13.00 & 12.88 & 10.75 & 11.53 & 11.60 & 11.18 & 11.4 & 11.35 & 11.28  \\
\textsc{Llama3.3 70B Instruct} & 19.50 & 12.70 & 15.78 & 17.38 & 17.60 & 17.45 & 17.82 & 18.60 & 19.00 \\ \midrule
\textsc{Qwen3 4B} & 8.40 & 8.25 & 8.30 & 9.83 & 8.18 & 7.67 & 8.65 & 9.15 & 7.88 &  \\
\textsc{Qwen3 30B A3B} & 23.70 & 18.93 & 21.18 & 19.50 & 21.12 & 22.10 & 21.20 & 21.22 & 22.15 \\
\textsc{Olmo 3 7B} & 5.40 & 5.12 & 5.15 & 4.52 & 5.33 & 5.40 & 5.08 & 5.85 & 5.75 \\
\textsc{Olmo 3 32B} &7.60 & 6.62 & 8.12 & 7.07 & 7.62 & 8.15 & 8.00 & 7.62 & 8.28  \\
\textsc{Olmo 3.1 32B} & 8.60 & 5.70 & 7.45 & 5.88 & 6.78 & 7.38 & 7.03 & 6.48 & 6.75 \\
\textsc{GPT-OSS 20B} & 22.30 & 13.70 & 17.68 & 11.80 & 20.70 & 18.50 & 17.80 & 14.57 & 17.75  \\
\textsc{GPT-OSS 120B} & 28.30 & 20.05 & 23.57 & 19.25 & 23.50 & 23.50 & 23.43 & 21.62 & 23.28 \\
\textsc{GPT-4.1} & 40.10 & 33.58 & 35.25 & 35.73 & 37.90 & 38.38 & 37.35 & 37.38 & 38.45 \\
\midrule
\\ \midrule 
& \textbf{ita-eng} & \textbf{jpn-eng} & \textbf{kor-eng} & \multicolumn{2}{c}{\textbf{mar-eng}}  & \multicolumn{2}{c}{\textbf{urd-eng}} & \textbf{zho-eng} & \textbf{avg.} \\
& & & & ori. & trans. & ori. & trans. & \\ \midrule
\textsc{Gemma3 12B} & 8.43 & 7.38  & 7.15 & 7.27 & 8.70 & 6.93 & 7.72 & 7.92 & 8.10 \\
\textsc{Gemma3 27B} & 11.68 & 11.20 & 11.00 & 10.20 & 11.85 & 11.12 & 11.22 & 11.85 & 11.38 \\
\textsc{Llama3.3 70B Instruct} & 18.10 & 16.83 & 17.88 & 16.70 & 17.55 & 17.32 & 17.72 & 16.88 & 17.21 \\ \midrule
\textsc{Qwen3 4B} & 8.22 & 7.90 & 7.88 & 7.52 & 8.55 & 8.20 & 8.40 & 9.43 & 8.38 \\
\textsc{Qwen3 30B A3B} & 21.35 & 21.10 & 21.12 & 21.15 & 19.00 & 21.62 & 20.40 & 20.28 & 20.84 \\
\textsc{Olmo 3 7B} & 5.50 & 5.40 & 4.88 & 4.65 & 4.47 & 4.90 & 5.35 & 5.17 & 5.16 \\
\textsc{Olmo 3 32B} & 8.30 & 7.90 & 7.50 & 7.58 & 6.28 & 7.38 & 7.15 & 7.60 & 7.57 \\
\textsc{Olmo 3.1 32B} & 6.20 & 7.35 & 6.40 & 6.78 & 5.55 & 6.28 & 6.70 & 6.78 & 6.59 \\
\textsc{GPT-OSS 20B} & 19.00 & 18.38 & 17.68 & 19.15 & 11.70 & 13.90 & 15.17 & 19.85 & 16.71\\
\textsc{GPT-OSS 120B} & 23.90 & 22.65 & 22.62 & 23.12 & 18.35 & 21.05 & 21.15 & 24.68 & 22.23\\
\textsc{GPT-4.1} & 38.35 & 36.78 & 36.12 & 35.68 & 35.05 & 36.00 & 36.65 & 36.20 & 36.55\\
\bottomrule
\end{tabular}
}
\caption{Fine-grained LLM results in F1 score averaged over all methods. \textbf{ori.} indicates the non transliteration. \textbf{trans.} indicates the transliteration.}
\label{tab:fine-grained-results}
\end{table*}

\paragraph{Generation.} We use \textsc{GPT-5.2} to produce perturbations and to transliterate text across all experimental settings.

\paragraph{Evaluation Judge.} We select \textsc{Qwen3-30B-A3B} as our automatic judge. While the original dataset~\cite{haas2025simpleqa} uses \textsc{GPT-4.1}, we find a high level of agreement between \textsc{Qwen3-30B-A3B} and \textsc{GPT-4.1}, with a Cohen’s $\kappa$ of 0.95. A full hyper-parameters of the models are reported in the Appendix~\ref{sec:hyper-parameters}.

\begin{figure*}[!th]
    \centering
    \includegraphics[width=\textwidth]{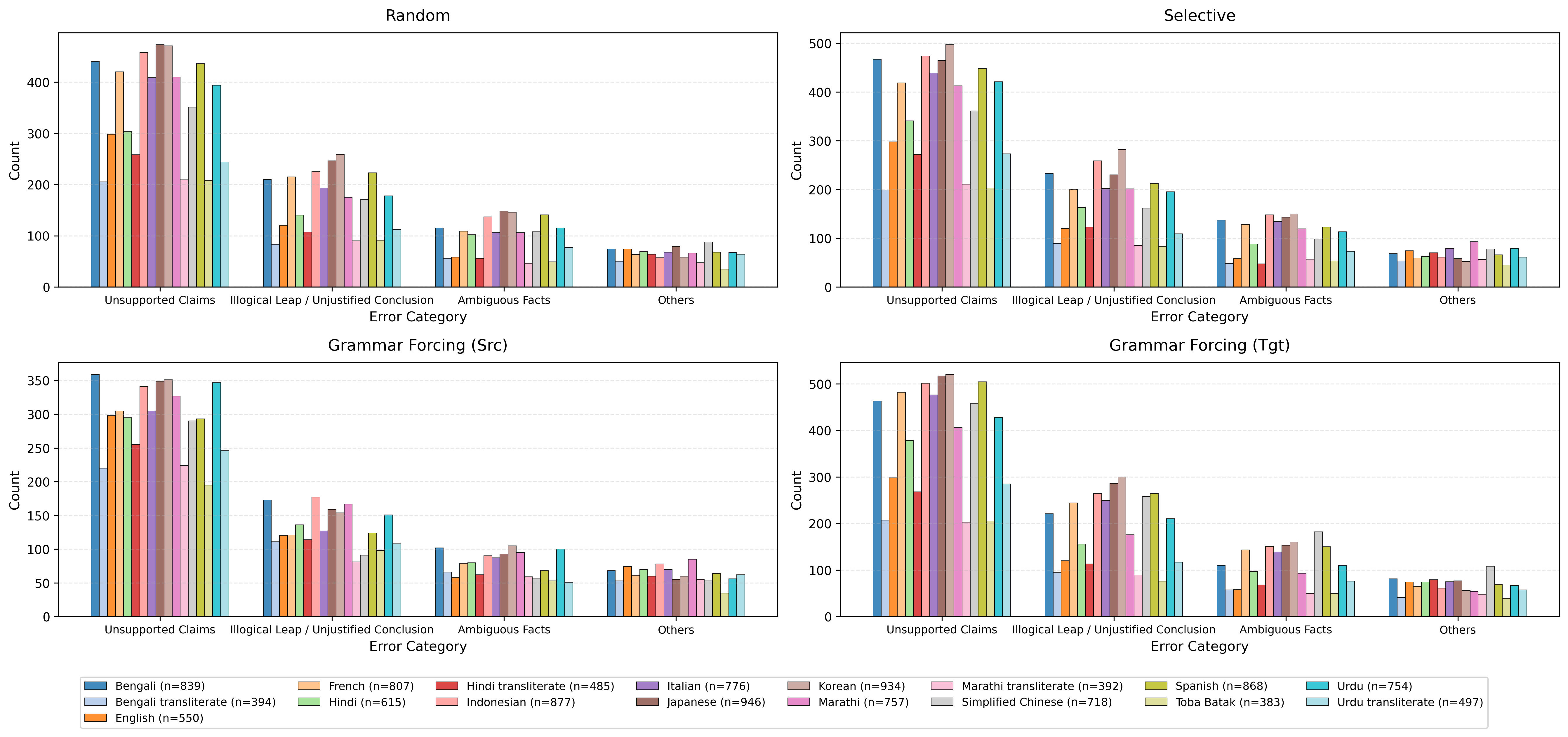}
    \caption{Reasoning error analysis of \textsc{Qwen3 30B A3B} using \textsc{GPT-OSS 120B} as LLM-as-a-judge across different subsets and languages of $\datasetname$. The analysis follows the error categorization of~\citet{ovalle2025beg}. We observe that errors are more frequent on non-English inputs, whereas transliterated text exhibits fewer errors.}
    \label{fig:reasoning}
\end{figure*}

\section{Results and Analysis}

In this section, we present and discuss the results of our experiments, offering a holistic analysis of LLMs’ abilities to understand, reason about, and generate code-switched text. We also examine the impact of transliteration on these capabilities in comparison to the original scripts.

\subsection{Overall Results}

\paragraph{Understanding.} Table~\ref{tab:results_perturbation} presents the performance of LLMs under different perturbation strategies. Overall, \textsc{Grammar Forcing (Src)} achieves the highest performance across all evaluated models, followed by \textsc{Selective} as the second-best strategy. On average, introducing perturbations results in an approximate 11\% performance drop across models. In contrast, \textsc{Grammar Forcing (Tgt)} generally underperforms relative to the other perturbation strategies. A more detailed, language-specific analysis is needed to determine which languages are better handled when the target language serves as the matrix language. In general, stronger models with reasoning capabilities perform better, and larger models in \textsc{Gemma3}, \textsc{Qwen3}, and \textsc{GPT-OSS} consistently achieve higher performance, with the exception of the \textsc{Olmo 3} models.

\paragraph{Reasoning.}
We investigate the reasoning behavior of all thinking models and find that their reasoning traces are consistently in English, even when the inputs are predominantly in non-English languages. This suggests that the models are highly English-centric, likely reflecting that their training data for thinking-oriented models is largely English, with minimal contributions from non-English data. We also observe instances of misalignment when perturbed text is passed to the LLM, resulting in degraded reasoning traces and negatively impacting overall performance.

As shown in Figure~\ref{fig:reasoning}, misalignment errors increase when code-switched text is used compared to the original English input. We conjecture that this is due to challenges in understanding and reasoning over mixed-language inputs. Notably, these errors are much less pronounced for transliterated text. Overall, misalignment is relatively evenly distributed across issue types, with the top three errors being \textit{Unsupported Claims}, \textit{Illogical Leap / Unjustified Conclusion}, and \textit{Ambiguous Facts}, following the taxonomy of~\citet{ovalle2025beg}.

\begin{table*}[!t]
\centering
\resizebox{\textwidth}{!}{
\begin{tabular}{lcccccccccccccccc|c}
\toprule
\textbf{Model} 
& \textbf{bbc-eng} & \multicolumn{2}{c}{\textbf{ben-eng}}  & \textbf{esp-eng} & \textbf{fra-eng} & \multicolumn{2}{c}{\textbf{hin-eng}} & \textbf{ind-eng} & \textbf{ita-eng} & \textbf{jpn-eng} & \textbf{kor-eng} & \multicolumn{2}{c}{\textbf{mar-eng}} & \multicolumn{2}{c}{\textbf{urd-eng}} & \textbf{zho-eng} & \textbf{avg.} \\ 
& & ori. & trans. & & & ori. & trans. & & & & & ori. & trans. & ori. & trans. & \\ \midrule
\textsc{Naturalness}
\\ \midrule
\textsc{Random} & 2.69 & 3.97 & 3.79 & 4.41 & 2.34 & 4.74 & 4.10 & 4.00 & 2.94 & 3.08 & 4.48 & 2.78 & 4.26 & 2.42 & 3.09 & 3.46 & 3.53\\
\textsc{Selective}  & \textbf{2.71} & 4.07 & 4.03 & 4.52 & 2.48 & 4.76 & 4.31 & 4.25 & 2.77 & 3.25 & 4.59 & 2.87 & 3.91 & \textbf{2.73} & \textbf{4.17} & 3.53 & 3.68 \\
\textsc{Grammar Force (src)}  & 2.47 & 2.85 & 3.05 & 4.02 & 1.09 & 4.38 & 3.89 & 3.70 & 2.17 & 2.84 & 4.06 & 2.22 & 3.14 & 2.35 & 2.47 & 3.08 & 2.99 \\
\textsc{Grammar Force (tgt)}  & 2.70 & \textbf{4.58} & \textbf{4.51} & \textbf{4.71} & \textbf{3.22} & \textbf{4.93} & \textbf{4.54} & \textbf{4.49} & \textbf{3.72} & \textbf{3.48} & \textbf{4.80} & \textbf{3.15} & \textbf{3.97} & 2.70 & 4.12 & \textbf{4.49} & \textbf{3.88} \\ \midrule
\textsc{Semantic Accuracy}
\\ \midrule
\textsc{Random} & 2.69 & \textbf{4.98} & \textbf{4.94} & 4.96 & \textbf{4.90} & \textbf{4.92} & 4.17 & 4.85 & 3.63 & 3.96 & \textbf{4.98} & 3.67 & 3.70 & 3.21 & 3.77 & \textbf{4.99} & 4.27 \\
\textsc{Selective}  & \textbf{2.71} & 4.87 & 4.90 & \textbf{4.99} & 4.85 & 4.88 & 4.46 & 4.93 & 3.53 & 3.94 & 4.97 & 3.73 & 3.99 & 3.47 & 3.23 & \textbf{4.99} & 4.22 \\
\textsc{Grammar Force (src)}  & 2.47 & \textbf{4.98} & \textbf{4.94} & 4.93 & 4.89 & 4.89 & 4.06 & 4.92 & 3.03 & 3.93 & \textbf{4.98} & 3.10 & 3.13 & 3.25 & 3.28 & 4.96 & 3.98 \\
\textsc{Grammar Force (tgt)}  & \textbf{2.71} & 4.96 & 4.91 & 4.98 & 4.86 & 4.88 & \textbf{4.67} & \textbf{4.94} & \textbf{4.36} & \textbf{3.97} & 4.96 & \textbf{3.82} & \textbf{4.03} & \textbf{3.50} & \textbf{4.66} & 4.97 & \textbf{4.45} \\
\bottomrule
\end{tabular}
}
\caption{Human evaluation results for code-switched texts generated using different strategies. Scores represent average ratings on a 5-point Likert scale, where higher values indicate better performance. Overall, \textsc{Grammar Force (Tgt)} outperforms other methods, suggesting that human annotators perceive code-switched texts as more natural and accurate when the syntactic structure follows the target language grammar.}
\label{tab:human-eval-results}
\end{table*}

\paragraph{Generation.}
Examining the human annotations, we find that restricting generation to target-language grammar via \textsc{Grammar Forcing (Tgt)} improves performance and outperforms all other perturbation strategies. In contrast, restricting generation to source-language grammar via \textsc{Grammar Forcing (Src)} yields the worst results. Interestingly, this observation is negatively correlated with the LLM understanding results: while \textsc{Grammar Forcing (Src)} performs well in comprehension evaluations, it fares poorly in naturalness according to human judgments. This suggests that LLMs may generally handle less perturbed non-English text better, highlighting a divergence between naturalness and understanding metrics.

In terms of naturalness, generations produced using the target-language grammar-forcing strategy are consistently rated higher than those from other approaches. Regarding accuracy, models largely preserve the original sentence meaning when generating code-switched outputs, particularly for language pairs involving higher-resource languages. The language pair with notably poor performance is \texttt{bbc-eng} (Toba Batak–English).

Annotation instructions and the evaluation rubric are provided in Appendix Section~\ref{sec:annotation-details}.


\subsection{Transliteration vs. Original Scripts}

In many languages, code-switching with English is expressed in romanized form rather than in the native script, often for ease of typing. In our dataset, we compare the transliterated and native-script versions of code-switching between English and Indian languages: Bengali-English (ben-eng), Hindi-English (hin-eng), Marathi-English (mar-eng), and Urdu-English (urd-eng).

\paragraph{Code-switched Question Understanding.} We compare all model's performance on scripted code-switched questions \textit{vs.} transliterations (see Table \ref{tab:fine-grained-results}). Across all four Indian languages, there is no clear pattern indicating that models understand one or the other better.

\paragraph{Code-switched Question Generation.} When comparing outputs in the original script with their transliterated counterparts, for Bengali–English code-switching, naturalness ratings are largely comparable across approaches (see Table \ref{tab:human-eval-results}). In contrast, for the Urdu–English pair and Marathi-English pair, transliterated sentences are rated as substantially more natural. For Hindi–English pairs, the original sentences are rated as slightly more natural. However, the annotators described that Hindi-English code-switching in roman script is dominant in informal communication, hence more natural (see qualitative analysis in Section \ref{app:qualitative-analysis} in Appendix). 
Overall, human annotators confirm that while transliteration sometimes leads to slightly less natural outputs due to the absence of standardized orthographic conventions and to more English-dominated grammar, they are much more representative to common code-switching writing practices.

\section{Conclusion}
We introduce $\datasetname$, a novel code-switching dataset with high-quality human annotations, covering 16 diverse parallel code-switched language pairs in both original scripts and transliterations. This dataset enables systematic evaluation of LLM performance across different code-switching patterns and text generation strategies. We analyze LLM reasoning behavior on code-switched question-answering tasks, revealing an average performance drop of approximately 11\% under code-switching conditions, while higher code-mixing density correlates with improved robustness, providing insights into how models process mixed-language inputs. Additionally, we conduct a systematic study of the naturalness and semantic accuracy of LLM-generated code-switched text, uncovering a tradeoff: outputs that preserve the non-English language as the matrix language are rated as more natural by native speakers, whereas English-dominant grammatical structures achieve higher downstream task performance but lower naturalness. In terms of reasoning, we find that code-switching perturbations negatively affect LLM reasoning traces, introducing unsupported claims, illogical leaps, and ambiguous facts that can lead to incorrect predictions. Together, these findings advance our understanding of LLM behavior in code-switching scenarios and provide a foundation for developing more robust, linguistically aware multilingual models. We hope that these insights will guide the design of future LLMs that are more resilient to language mixing.

\section*{Limitations}
We acknowledge certain model limitations: due to budget constraints, we are unable to evaluate all possible models. Additionally, we leverage an AI assistant to paraphrase and improve our writing, as well as to identify typographical and stylistic issues. We also acknowledge the source of our dataset; it was selected based on its difficulty at the time of the dataset and paper creation, ensuring that it continues to pose meaningful challenges for LLMs, including proprietary models.

\section*{Acknowledgments}
We would like to sincerely thank Aoyama Amane and Hiroto Aoyama for their valuable assistance during the human annotation study.

\bibliography{custom}

\clearpage
\appendix


\section{Pre-processing for Evaluation}
\label{sec:preprocessing}

Pre‑processing is conducted with careful consideration of the structural diversity across the languages represented in $\datasetname$. Several languages in the dataset, such as Japanese, Korean, and Chinese, do not use whitespace as a reliable indicator of word boundaries, while others, including Hindi, Urdu, Bengali, and Marathi, display orthographic and morphological complexity that makes naive whitespace segmentation insufficient. To ensure a consistent linguistic representation across languages and scripts, we adopt language‑specific tokenization strategies aligned with the typological properties of each language.

For Indic languages (Hindi, Urdu, Bengali, and Marathi), we use the iNLTK tokenizer~\citep{arora2020inltk},\footnote{\scriptsize\url{https://inltk.readthedocs.io/}} which provides script‑aware segmentation tailored to the Devanagari, Perso‑Arabic, and Eastern Brahmic writing systems. Japanese text is processed using the Tohoku NLP Group's \texttt{bert-japanese}\footnote{\scriptsize\url{https://hf.co/tohoku-nlp/bert-base-japanese-v2}} tokenizer, which performs word‑level segmentation suitable for Japanese’s inflectional morphology. Korean text is tokenized using the KLUE BERT~\citep{park2021klue}\footnote{\scriptsize\url{https://hf.co/klue/bert-base}} tokenizer, which applies morpheme‑based segmentation to better reflect the agglutinative structure of Korean. For Simplified Chinese, we apply the BERT‑base Chinese~\citep{devlin2019bert}\footnote{\scriptsize\url{https://hf.co/google-bert/bert-base-chinese}} tokenizer, which implements the standard character‑based segmentation widely used in Chinese NLP.

Because $\datasetname$ includes both original‑script and transliterated variants for several language pairs, we handle transliteration in a controlled and parallel manner. Original‑script and transliterated forms are processed using aligned tokenization procedures to minimize discrepancies arising from preprocessing rather than linguistic content. This ensures that comparisons across script conditions capture genuine variation in model behavior rather than artifacts introduced by segmentation.

Together, these preprocessing steps provide a consistent linguistic foundation for the human annotation workflow and the downstream evaluation of model performance. By standardizing tokenization across diverse scripts and transliteration settings, this stage reduces structural variation unrelated to model capability and supports more reliable analysis within the broader methodological pipeline.

\section{Detailed Results}
Tables~\ref{tab:random-fine-grained-results},~\ref{tab:selective-fine-grained-results},~\ref{tab:grammar-forcing-source-fine-grained-results},~\ref{tab:grammar-forcing-target-fine-grained-results} present the fine-grained results for the \textsc{Random}, \textsc{Selective}, \textsc{Grammar Forcing (Src)}, and \textsc{Grammar Forcing (Tgt)} settings, respectively.

\begin{table*}[!th]
\centering
\resizebox{\textwidth}{!}{
\begin{tabular}{lccccccccccccc}
\toprule
\textbf{Model} & \textbf{No Perturb.} & \textbf{bbc-eng} & \multicolumn{2}{c}{\textbf{ben-eng}} & \textbf{esp-eng} & \textbf{fra-eng} & \multicolumn{2}{c}{\textbf{hin-eng}} & \textbf{ind-eng} \\
& & & ori. & trans. & & & ori. & trans. & \\ \midrule
\textsc{Gemma3 12B} & 8.9 & 12.0 & 7.3 & 10.2 & 7.5 & 7.8 & 7.7 & 8.6 & 8.1 \\
\textsc{Gemma3 27B} & 13.0 & 12.6 & 11.9 & 9.5 & 12.0 & 11.1 & 11.3 & 11.4 & 11.0 \\
\textsc{Llama3.3 70B Instruct} & 19.5 & 11.1 & 15.0 & 16.8 & 17.1 & 18.0 & 18.5 & 19.1 & 19.9 \\ \midrule
\textsc{Qwen3 4B} & 8.4 & 8.8 & 7.8 & 9.9 & 7.9 & 7.6 & 9.2 & 9.4 & 7.7 \\
\textsc{Qwen3 30B A3B} & 23.7 & 19.0 & 21.4 & 19.2 & 20.5 & 22.0 & 21.4 & 21.6 & 21.3 \\
\textsc{Olmo 3 7B} & 5.4 & 5.0 & 5.3 & 5.0 & 5.3 & 5.7 & 5.1 & 6.4 & 6.4 \\
\textsc{Olmo 3 32B} & 7.6 & 5.9 & 7.9 & 7.4 & 7.2 & 6.8 & 8.3 & 7.8 & 8.3 \\
\textsc{Olmo 3.1 32B} & 8.6 & 4.7 & 7.9 & 4.6 & 6.4 & 7.7 & 7.0 & 6.1 & 7.2 \\
\textsc{GPT-OSS 20B} & 22.3 & 13.4 & 17.7 & 12.3 & 19.2 & 18.1 & 18.0 & 15.1 & 17.8 \\
\textsc{GPT-OSS 120B} & 28.3 & 19.6 & 23.8 & 20.5 & 23.1 & 23.1 & 20.7 & 21.9 & 23.0 \\
\textsc{GPT-4.1} & 40.1 & 33.7 & 35.1 & 35.5 & 37.3 & 38.5 & 37.9 & 37.8 & 40.3 \\
\midrule
\\ \midrule
& \textbf{ita-eng} & \textbf{jpn-eng} & \textbf{kor-eng} & \multicolumn{2}{c}{\textbf{mar-eng}} & \multicolumn{2}{c}{\textbf{urd-eng}} & \textbf{zho-eng} & \textbf{avg.} \\
& & & & ori. & trans. & ori. & trans. & \\ \midrule
\textsc{Gemma3 12B} & 8.5 & 7.5 & 6.6 & 7.5 & 8.3 & 7.0 & 7.8 & 7.6 & 8.1 \\
\textsc{Gemma3 27B} & 13.6 & 11.2 & 10.8 & 10.0 & 12.4 & 11.5 & 11.8 & 11.9 & 11.5 \\
\textsc{Llama3.3 70B Instruct} & 17.4 & 17.0 & 17.7 & 16.5 & 17.8 & 16.1 & 17.0 & 16.9 & 17.0 \\ \midrule
\textsc{Qwen3 4B} & 7.9 & 7.7 & 6.8 & 6.4 & 8.4 & 8.3 & 9.1 & 9.0 & 8.2 \\
\textsc{Qwen3 30B A3B} & 20.9 & 21.9 & 21.2 & 20.5 & 20.3 & 21.4 & 20.4 & 21.3 & 20.9 \\
\textsc{Olmo 3 7B} & 5.4 & 5.5 & 5.1 & 4.6 & 5.1 & 4.2 & 5.8 & 4.8 & 5.3 \\
\textsc{Olmo 3 32B} & 8.1 & 8.0 & 7.2 & 7.3 & 6.1 & 7.8 & 7.1 & 7.8 & 7.4 \\
\textsc{Olmo 3.1 32B} & 6.0 & 6.6 & 6.4 & 6.9 & 5.6 & 6.6 & 6.5 & 6.9 & 6.4 \\
\textsc{GPT-OSS 20B} & 19.1 & 16.9 & 16.4 & 19.1 & 10.3 & 15.1 & 14.3 & 18.5 & 16.3 \\
\textsc{GPT-OSS 120B} & 22.5 & 21.1 & 22.0 & 22.2 & 16.8 & 21.0 & 20.4 & 23.8 & 21.6 \\
\textsc{GPT-4.1} & 37.5 & 36.9 & 35.4 & 35.4 & 35.0 & 36.2 & 36.3 & 36.0 & 36.5 \\
\bottomrule
\end{tabular}
}
\caption{Fine-grained LLM results in accuracy on \textsc{Random}. \textbf{ori.} indicates the non transliteration. \textbf{trans.} indicates the transliteration.}
\label{tab:random-fine-grained-results}
\end{table*}

\begin{table*}[!th]
\centering
\resizebox{\textwidth}{!}{
\begin{tabular}{lccccccccccccc}
\toprule
\textbf{Model} & \textbf{No Perturb.} & \textbf{bbc-eng} & \multicolumn{2}{c}{\textbf{ben-eng}} & \textbf{esp-eng} & \textbf{fra-eng} & \multicolumn{2}{c}{\textbf{hin-eng}} & \textbf{ind-eng} \\
& & & ori. & trans. & & & ori. & trans. & \\ \midrule
\textsc{Gemma3 12B} & 8.9 & 10.7 & 7.3 & 10.8 & 7.9 & 7.4 & 9.0 & 8.0 & 8.6 \\
\textsc{Gemma3 27B} & 13.0 & 13.1 & 10.0 & 12.9 & 12.9 & 11.4 & 11.9 & 11.2 & 12.1 \\
\textsc{Llama3.3 70B Instruct} & 19.5 & 13.2 & 16.7 & 18.6 & 17.6 & 17.3 & 19.0 & 19.0 & 19.4 \\
\midrule
\textsc{Qwen3 4B} & 8.4 & 8.4 & 8.3 & 10.2 & 8.6 & 6.2 & 7.6 & 7.9 & 8.3 \\
\textsc{Qwen3 30B A3B} & 23.7 & 17.9 & 20.2 & 20.3 & 20.9 & 22.5 & 21.2 & 20.0 & 22.8 \\
\textsc{Olmo 3 7B} & 5.4 & 5.1 & 5.9 & 5.0 & 5.3 & 5.2 & 5.5 & 6.0 & 5.2 \\
\textsc{Olmo 3 32B} & 7.6 & 7.3 & 8.8 & 7.1 & 8.1 & 7.6 & 7.3 & 7.0 & 9.1 \\
\textsc{Olmo 3.1 32B} & 8.6 & 5.0 & 7.7 & 5.8 & 7.3 & 7.1 & 7.3 & 6.4 & 6.3 \\
\textsc{GPT-OSS 20B} & 22.3 & 13.5 & 17.0 & 10.9 & 20.6 & 16.6 & 18.0 & 13.3 & 16.4 \\
\textsc{GPT-OSS 120B} & 28.3 & 20.7 & 24.1 & 18.6 & 22.9 & 21.8 & 24.4 & 22.2 & 23.5 \\
\textsc{GPT-4.1} & 40.1 & 34.1 & 34.6 & 36.7 & 37.9 & 38.1 & 37.2 & 37.5 & 37.2 \\
\midrule
\\ \midrule
& \textbf{ita-eng} & \textbf{jpn-eng} & \textbf{kor-eng} & \multicolumn{2}{c}{\textbf{mar-eng}} & \multicolumn{2}{c}{\textbf{urd-eng}} & \textbf{zho-eng} & \textbf{avg.} \\
& & & & ori. & trans. & ori. & trans. & \\ \midrule
\textsc{Gemma3 12B} & 7.5 & 7.5 & 7.3 & 7.2 & 9.5 & 6.5 & 8.1 & 8.2 & 8.2 \\
\textsc{Gemma3 27B} & 11.1 & 11.5 & 11.3 & 9.9 & 11.1 & 10.5 & 10.7 & 12.0 & 11.5 \\
\textsc{Llama3.3 70B Instruct} & 19.1 & 17.6 & 18.6 & 17.0 & 17.7 & 19.2 & 18.4 & 16.6 & 17.8 \\
\midrule
\textsc{Qwen3 4B} & 8.1 & 7.5 & 9.5 & 7.4 & 8.0 & 7.7 & 8.5 & 9.2 & 8.2 \\
\textsc{Qwen3 30B A3B} & 20.3 & 21.0 & 19.8 & 20.9 & 18.5 & 20.8 & 21.1 & 21.0 & 20.6 \\
\textsc{Olmo 3 7B} & 5.5 & 5.7 & 5.1 & 4.5 & 4.2 & 5.6 & 5.8 & 5.2 & 5.3 \\
\textsc{Olmo 3 32B} & 8.0 & 7.5 & 6.6 & 7.6 & 7.0 & 6.7 & 7.8 & 7.8 & 7.6 \\
\textsc{Olmo 3.1 32B} & 6.4 & 7.5 & 7.0 & 7.7 & 6.2 & 6.4 & 7.9 & 6.2 & 6.8 \\
\textsc{GPT-OSS 20B} & 17.9 & 17.6 & 16.0 & 19.2 & 11.1 & 12.5 & 15.4 & 19.0 & 15.9 \\
\textsc{GPT-OSS 120B} & 23.5 & 23.1 & 22.1 & 23.0 & 19.5 & 21.1 & 20.8 & 23.4 & 22.2 \\
\textsc{GPT-4.1} & 38.3 & 36.9 & 35.1 & 34.9 & 34.9 & 35.8 & 37.3 & 35.6 & 36.4 \\
\bottomrule
\end{tabular}
}
\caption{Fine-grained LLM results in accuracy on \textsc{Selective}. \textbf{ori.} indicates the non transliteration. \textbf{trans.} indicates the transliteration.}
\label{tab:selective-fine-grained-results}
\end{table*}

\begin{table*}[!th]
\centering
\resizebox{\textwidth}{!}{
\begin{tabular}{lccccccccccccc}
\toprule
\textbf{Model} & \textbf{No Perturb.} & \textbf{bbc-eng} & \multicolumn{2}{c}{\textbf{ben-eng}} & \textbf{esp-eng} & \textbf{fra-eng} & \multicolumn{2}{c}{\textbf{hin-eng}} & \textbf{ind-eng} \\
& & & ori. & trans. & & & ori. & trans. & \\ \midrule
\textsc{Gemma3 12B} & 8.9 & 10.2 & 8.0 & 8.3 & 7.8 & 7.3 & 8.5 & 7.5 & 8.5 \\
\textsc{Gemma3 27B} & 13.0 & 12.5 & 11.7 & 12.4 & 10.7 & 11.0 & 10.8 & 11.2 & 11.2 \\
\textsc{Llama3.3 70B Instruct} & 19.5 & 13.5 & 17.0 & 14.9 & 16.8 & 16.7 & 16.1 & 17.5 & 17.8 \\ \midrule
\textsc{Qwen3 4B} & 8.4 & 8.8 & 9.1 & 11.2 & 7.8 & 8.9 & 9.1 & 10.5 & 7.5 \\
\textsc{Qwen3 30B A3B} & 23.7 & 21.3 & 23.0 & 20.4 & 22.0 & 22.0 & 21.8 & 20.8 & 22.6 \\
\textsc{Olmo 3 7B} & 5.4 & 5.1 & 4.8 & 4.3 & 4.9 & 4.7 & 5.0 & 6.0 & 5.4 \\
\textsc{Olmo 3 32B} & 7.6 & 7.1 & 7.5 & 6.6 & 7.1 & 9.5 & 8.0 & 7.1 & 7.5 \\
\textsc{Olmo 3.1 32B} & 8.6 & 6.9 & 7.3 & 6.1 & 7.0 & 7.2 & 7.1 & 7.0 & 7.1 \\
\textsc{GPT-OSS 20B} & 22.3 & 14.5 & 18.3 & 15.0 & 22.1 & 19.8 & 16.9 & 14.8 & 18.6 \\
\textsc{GPT-OSS 120B} & 28.3 & 20.6 & 24.2 & 21.0 & 23.8 & 24.8 & 23.3 & 21.9 & 24.0 \\
\textsc{GPT-4.1} & 40.1 & 33.7 & 35.6 & 36.2 & 37.8 & 38.0 & 35.7 & 38.2 & 38.2 \\
\midrule
\\ \midrule
& \textbf{ita-eng} & \textbf{jpn-eng} & \textbf{kor-eng} & \multicolumn{2}{c}{\textbf{mar-eng}} & \multicolumn{2}{c}{\textbf{urd-eng}} & \textbf{zho-eng} & \textbf{avg.} \\
& & & & ori. & trans. & ori. & trans. & \\ \midrule
\textsc{Gemma3 12B} & 9.4 & 7.7 & 7.8 & 7.4 & 8.2 & 7.7 & 7.5 & 8.3 & 8.1 \\
\textsc{Gemma3 27B} & 11.3 & 11.3 & 10.7 & 10.5 & 12.5 & 11.5 & 11.0 & 11.6 & 11.4 \\
\textsc{Llama3.3 70B Instruct} & 17.4 & 14.5 & 17.4 & 16.4 & 16.0 & 16.1 & 17.1 & 16.4 & 16.4 \\ \midrule
\textsc{Qwen3 4B} & 9.8 & 8.0 & 7.8 & 8.3 & 9.4 & 8.2 & 8.8 & 9.0 & 8.9 \\
\textsc{Qwen3 30B A3B} & 21.8 & 20.0 & 21.1 & 21.9 & 20.8 & 22.6 & 20.4 & 20.7 & 21.4 \\
\textsc{Olmo 3 7B} & 5.8 & 4.8 & 4.2 & 4.6 & 4.3 & 4.9 & 4.8 & 4.6 & 4.9 \\
\textsc{Olmo 3 32B} & 8.8 & 8.0 & 7.7 & 7.0 & 6.9 & 7.6 & 6.6 & 6.9 & 7.5 \\
\textsc{Olmo 3.1 32B} & 6.3 & 7.7 & 5.3 & 6.5 & 5.7 & 5.9 & 6.6 & 5.6 & 6.6 \\
\textsc{GPT-OSS 20B} & 20.8 & 26.0 & 22.4 & 20.1 & 13.1 & 17.5 & 17.6 & 22.0 & 18.7 \\
\textsc{GPT-OSS 120B} & 25.6 & 28.5 & 26.3 & 26.0 & 20.4 & 22.9 & 23.3 & 28.7 & 24.1 \\
\textsc{GPT-4.1} & 39.5 & 37.4 & 38.7 & 37.2 & 35.6 & 36.5 & 36.5 & 36.6 & 37.0 \\
\bottomrule
\end{tabular}
}
\caption{Fine-grained LLM results in accuracy on \textsc{Grammar Forcing (Src)}. \textbf{ori.} indicates the non transliteration. \textbf{trans.} indicates the transliteration.}
\label{tab:grammar-forcing-source-fine-grained-results}
\end{table*}

\begin{table*}[!th]
\centering
\resizebox{\textwidth}{!}{
\begin{tabular}{lccccccccccccc}
\toprule
\textbf{Model} & \textbf{No Perturb.} & \textbf{bbc-eng} & \multicolumn{2}{c}{\textbf{ben-eng}} & \textbf{esp-eng} & \textbf{fra-eng} & \multicolumn{2}{c}{\textbf{hin-eng}} & \textbf{ind-eng} \\
& & & ori. & trans. & & & ori. & trans. & \\ \midrule
\textsc{Gemma3 12B} & 8.9 & 11.5 & 7.0 & 9.5 & 7.3 & 7.6 & 8.0 & 8.2 & 8.0 \\
\textsc{Gemma3 27B} & 13.0 & 13.3 & 9.4 & 11.3 & 10.8 & 11.2 & 11.6 & 11.6 & 10.8 \\
\textsc{Llama3.3 70B Instruct} & 19.5 & 13.0 & 14.4 & 19.2 & 18.9 & 17.8 & 17.7 & 18.8 & 18.9 \\ \midrule
\textsc{Qwen3 4B} & 8.4 & 7.0 & 8.0 & 8.0 & 8.4 & 8.0 & 8.7 & 8.8 & 8.0 \\
\textsc{Qwen3 30B A3B} & 23.7 & 17.5 & 20.1 & 18.1 & 21.1 & 21.9 & 20.4 & 22.5 & 21.9 \\
\textsc{Olmo 3 7B} & 5.4 & 5.3 & 4.6 & 3.8 & 5.8 & 6.0 & 4.7 & 5.0 & 6.0 \\
\textsc{Olmo 3 32B} & 7.6 & 6.2 & 8.3 & 7.2 & 8.1 & 8.7 & 8.4 & 8.6 & 8.2 \\
\textsc{Olmo 3.1 32B} & 8.6 & 6.2 & 6.9 & 7.0 & 6.4 & 7.5 & 6.7 & 6.4 & 6.4 \\
\textsc{GPT-OSS 20B} & 22.3 & 13.4 & 17.7 & 9.0 & 20.9 & 19.5 & 18.3 & 15.1 & 18.2 \\
\textsc{GPT-OSS 120B} & 28.3 & 19.3 & 22.2 & 16.9 & 24.2 & 24.3 & 25.3 & 20.5 & 22.6 \\
\textsc{GPT-4.1} & 40.1 & 32.8 & 35.7 & 34.5 & 38.6 & 38.9 & 38.6 & 36.0 & 38.1 \\
\midrule
\\ \midrule
& \textbf{ita-eng} & \textbf{jpn-eng} & \textbf{kor-eng} & \multicolumn{2}{c}{\textbf{mar-eng}} & \multicolumn{2}{c}{\textbf{urd-eng}} & \textbf{zho-eng} & \textbf{avg.} \\
& & & & ori. & trans. & ori. & trans. & \\ \midrule
\textsc{Gemma3 12B} & 8.3 & 6.8 & 6.9 & 7.0 & 8.8 & 6.5 & 7.5 & 7.6 & 7.9 \\
\textsc{Gemma3 27B} & 10.7 & 10.8 & 11.2 & 10.4 & 11.4 & 11.0 & 11.4 & 11.9 & 11.2 \\
\textsc{Llama3.3 70B Instruct} & 18.5 & 18.2 & 17.8 & 16.9 & 18.7 & 17.9 & 18.4 & 17.6 & 17.7 \\ \midrule
\textsc{Qwen3 4B} & 7.1 & 8.4 & 7.4 & 8.0 & 8.4 & 8.6 & 7.2 & 10.5 & 8.2 \\
\textsc{Qwen3 30B A3B} & 22.4 & 21.5 & 22.4 & 21.3 & 16.4 & 21.7 & 19.7 & 18.1 & 20.4 \\
\textsc{Olmo 3 7B} & 5.3 & 5.6 & 5.1 & 4.9 & 4.3 & 4.9 & 5.0 & 6.1 & 5.2 \\
\textsc{Olmo 3 32B} & 8.3 & 8.1 & 8.5 & 8.4 & 5.1 & 7.4 & 7.1 & 7.9 & 7.8 \\
\textsc{Olmo 3.1 32B} & 6.1 & 7.6 & 6.9 & 6.0 & 4.7 & 6.2 & 5.8 & 8.4 & 6.6 \\
\textsc{GPT-OSS 20B} & 18.2 & 13.0 & 15.9 & 18.2 & 12.3 & 10.5 & 13.4 & 19.9 & 15.8 \\
\textsc{GPT-OSS 120B} & 24.0 & 17.9 & 20.1 & 21.3 & 16.7 & 19.2 & 20.1 & 22.8 & 21.1 \\
\textsc{GPT-4.1} & 38.1 & 35.9 & 35.3 & 35.2 & 34.7 & 35.5 & 36.5 & 36.6 & 36.3 \\
\bottomrule
\end{tabular}
}
\caption{Fine-grained LLM results in accuracy on \textsc{Grammar Forcing (Tgt)}. \textbf{ori.} indicates the non transliteration. \textbf{trans.} indicates the transliteration.}
\label{tab:grammar-forcing-target-fine-grained-results}
\end{table*}

\section{Annotation Details}
\label{sec:annotation-details}

\subsection{Instructions and Annotation Rubrics}
\label{sec:annotation-rubrics}
Table~\ref{tab:rating-rubric} describes the rubric given to the annotators to evaluate the quality of the code-switched questions along two axes: \textit{Naturalness} -- how understandable and natural the sentence is for a bilingual speaker -- and \textit{Accuracy} -- how accurate the code-switched sentence is compared to the original English one. All annotators were kept blind to the models involved: they did not have access to labels, system outputs, or information regarding the methods used to generate the texts they annotated, reducing potential bias during the annotation process.

\begin{table*}[t]
\centering
\small
\begin{tabular}{c|p{0.42\textwidth}|p{0.42\textwidth}}
\toprule
\textbf{Rating} & \textbf{Naturalness (Fluency/Readability)} & \textbf{Accuracy (Fidelity/Correctness)} \\
\midrule
5 & Excellent: Completely natural and fluent. Reads as if originally written by a native speaker. Zero awkward phrasing, unidiomatic expressions, or grammatical errors. & Excellent: Flawlessly accurate. All key information, facts, names, and numerical values are perfectly conveyed and correct according to the source. \\
\midrule
4 & Good: Highly natural and fluent. Reads very well, with only a few minor, almost unnoticeable awkward turns of phrase or non-native sounding expressions that do not hinder understanding. & Good: Mostly accurate. Only minor, non-critical details or nuances may be slightly off, but the core meaning, all essential facts, and numerical values are correctly represented. \\
\midrule
3 & Average: Understandable but noticeable unnatural. Contains several clear examples of awkward wording, stilted phrasing, or minor grammatical issues that distract the reader but do not prevent comprehension. & Average: Generally accurate. The main idea is conveyed, but there are minor factual errors, slightly incorrect names, or omissions that require minor mental correction. \\
\midrule
2 & Poor: Significantly unnatural, halting, or confusing. Requires significant effort to read and interpret due to numerous awkward structures, poor word choices, or frequent major grammatical errors. & Poor: Contains significant errors. Core facts, names, or numerical values are noticeably incorrect, or major parts of the information are missing, substantially distorting the original meaning. \\
\midrule
1 & Very Poor: Completely unnatural, garbled, or incoherent. Virtually unintelligible or reads like a word-for-word machine translation. Meaning cannot be reliably discerned. & Very Poor: Fundamentally inaccurate. The output is largely irrelevant, contains critical, misleading factual errors, or fails to address the original prompt/source content effectively. \\
\bottomrule
\end{tabular}
\caption{Rating rubric for evaluating translation quality across naturalness and accuracy dimensions.}
\label{tab:rating-rubric}
\end{table*}

\subsection{Recruitment and Demography}




To ensure a diverse yet methodologically consistent annotator pool, contributors were grouped into two age‑based demographic categories: Cat-1 (13–28 years) and Cat-2 (29–44 years). These categories were selected to capture variation between younger and more mature adult speakers while maintaining sufficient sample sizes within each group.

The study employed annotators representing a broad range of linguistic backgrounds. The final distribution was as follows: Indonesian (1 annotator: Cat-1), Japanese (4 annotators: 1 Cat-2 and 3 Cat-1), Bengali (1: Cat-2), French (2: 1 Cat-1 and 1 Cat-2), Spanish (1: Cat-2), Chinese (1: Cat-2), Marathi (2: both Cat-2), Hindi (1: Cat-2), Urdu (1: Cat-1), Toba Batak (1: Cat-1), Italian (1: Cat-2), and Korean (1: Cat-1).
With respect to language expertise, all annotators are native speakers of the target language they worked on, with the exception of the Japanese group: one annotator is a non‑native speaker who has lived in Japan for over eight years and is fluent in Japanese, while two native speakers were raised in Southeast-Asia and attended international schools.

In terms of educational background, except for one Indonesian annotator, all annotators had completed a university degree, and the majority held a master’s degree or higher. Many are actively involved in research in computational linguistics or adjacent fields, ensuring familiarity with linguistic concepts relevant to annotation tasks.

Annotators were recruited based on two principal criteria: demonstrated fluency in both English and the target language, and willingness to contribute to research‑oriented linguistic annotation. No additional professional prerequisites were imposed. Additionally, we receive the consent from all annotators to release the dataset as open-source.

\section{Hyper-parameters}
\label{sec:hyper-parameters}

For inference, we use 4 H100 80GB GPUs with maximum of 16,384 tokens using vLLM~\citep{kwon2023efficient}. We configured generation hyper-parameters separately for each model family based on the models’ default decoding settings. For \textsc{Gemma3 12B} and \textsc{Gemma3 27B}, we used a temperature of 1.0 with nucleus sampling set to top-$p = 0.95$ and top-$k = 64$. For \textsc{GPT-OSS 120B} and \textsc{GPT-OSS 20B}, generation was performed with a temperature of 1.0. For Llama-3.3-70B Instruct, we used temperature of 0.6 with nucleus sampling set to top-$p = 0.9$. For \textsc{OLMo 3 7B}, \textsc{Olmo 3 32B}, and \textsc{Olmo 3.1 32B}, we set the temperature to 0.6 and used nucleus sampling with top-$p = 0.95$. For \textsc{Qwen 3 30B A3B} and \textsc{Qwen3 4B}, we used a temperature of 0.6, nucleus sampling with top-$p = 0.95$, and top-$k = 20$. For perturbation, we use low reasoning level for \textsc{GPT-5.2}.

\section{Qualitative Analysis}\label{app:qualitative-analysis}

\paragraph{Spanish.}
The annotators observe high accuracy scores ($\geq$4) across code-switched Spanish variants, with only a small number of failures (N < 5) stemming from instruction-following or formatting errors. Interestingly, naturalness scores varied with the form of code-mixing rather than its amount. 
They observe lower naturalness ($\leq$3) in cases of \textit{grammatical tension}, where English and Spanish introduced incompatible sentence structures within the same clause, such as English question framing appearing alongside Spanish verb forms or word order (e.g., ‘In which año did', 'Contra quién did'). 
In contrast, higher naturalness scores ($\geq$4) occurred when one language determined the clause-level sentence structure from start to finish, either Spanish throughout (e.g., 'Cuánto dinero, en euros, se ordenó al cirujano?') or English throughout with Spanish appearing only as inserted words (e.g., 'What posición was Everton in when Rafael Benítez was sacked...'). Overall, human annotators found high naturalness aligned with consistency in clause-level sentence structure rather than with the presence or amount of lexical mixing.

\paragraph{Korean.}
In general, Korean--English code-switching outputs are both natural and accurate, consistently scoring above 4 out of 5 across all settings.
Notably, the outputs are nearly error-free in terms of semantic correctness, achieving an accuracy of 4.97 out of 5 on average.

Despite the high quality, the annotators observe recurring minor issues that slightly reduce perceived naturalness.
First, all perturbations result in English-centric conventions even when Korean is the majority language of the sentence, particularly in format-sensitive slots such as dates and numbers.
For example, dates are often rendered in a day--month--year order (``날짜, 월, 연도''), whereas Korean typically prefers year--month--day; this tendency is evident in queries such as:
``\emph{Holland의 William II는 Germany의 King이 된 날짜, 월, 연도가 언제였나요?}''

Second, human annotators observe unnatural switch points that occur inside proper-name phrases or tight noun compounds, yielding mixed phrases like `\emph{patch 노트} (patch note)', `\emph{뱀파이어 number} (vampire number)', and `\emph{recreational 수학} (recreational mathematics)'.

Third, in English-dominant sentences, the model sometimes inserts Korean loanwords (\emph{i.e.}, phonetic transliterations of English using Korean scripts) as embedded items. This can sound marked in otherwise such sentences.
For example, the following query contains multiple such insertions:
``\emph{Which DropoutTV 시리즈 is a 스핀오프 of `Game Changer' inspired by its `Noise Boys' 에피소드들?}'', where all Korean embedded phrases (\emph{i.e.}, `\emph{시리즈 (series)}', `\emph{스핀오프 (spin-off)}', `\emph{에피소드 (episode)}') are loanwords from English.
While the intended meaning remains clear, these insertions can be perceived as stylistically inconsistent given the surrounding English context.

\paragraph{Chinese.}
For semantic accuracy, in a small number of cases, models introduce extraneous content by continuing to answer the input prompt rather than strictly transforming it into a code-switched sentence. Nevertheless, the majority of generated outputs preserve the original meaning and remain semantically faithful to the source sentences.

In terms of naturalness, models are generally able to produce natural-sounding Chinese–English code-switched sentences for inputs with relatively simple syntactic structures. As syntactic complexity increases, however, switch points become more arbitrary and less linguistically natural. In addition, human annotators consistently prefer Chinese-dominated outputs rather than English-dominated.

\paragraph{Japanese.}
The model generally produces natural, accurate Japanese-English code-switched sentences. 
Human annotators observe that code-switched segments can be written in either Japanese script (Katakana), typically used to express non-Japanese words, or in the Latin script.
In terms of naturalness, while sentences where nouns or noun phrases are used in one language with the syntax of the other language are perceived as more natural, those where verbs or verb phrases are used in one language with the syntax of the other language are less natural.
Since Japanese and English differ in writing systems, word order, and the expression of grammatical categories, the use of code-switching is inherently limited, which in turn affects the naturalness judgments.
Regardless of the annotators' first language, sentences with Japanese syntax are judged to be more natural on average than those with English syntax.
For example, one of the code-switched sentences of ``On what day, month, and year was Algerian artist Mohammed Racim born?'' is ``\textit{アルジェリア}の芸術家\textit{モハメッド・ラシム}は、何年何月何日に生まれましたか？'' (Katakana is in italic).
Such sentences with Katakana are frequently observed in Japanese utterances and are therefore considered more natural than those that mix Japanese and English phrases.

\paragraph{Hindi.}
Human annotators observe the model generates fluent and contextually appropriate Hindi–English mixed outputs for a given English input, indicating strong code-switching capabilities. However, a mismatch emerges between model outputs and real-world informal usage. In casual digital communication, Hindi–English mixing is predominantly written in the Latin script, whereas many model generations include Devanagari script within mixed text. Although these outputs are linguistically accurate and natural, they do not reflect common informal writing practices. 
While Hinglish reflects informal spoken interaction among Hindi speakers, it is less characteristic of formal discourse. Overall, annotators do not identify a consistent pattern in which an LLM reliably generates code-switched text exclusively in the Latin script, despite this being dominant in informal communication. Additionally, many model-generated outputs are stylistically more refined than the original English inputs, suggesting that LLMs tend to enhance fluency and polish even when generating code-switched text.

\paragraph{Marathi.}

Marathi outputs in the native Devanagari script are generally well-formed and fluent, aligning closely with standard written Marathi. Naturalness is generally high, reflecting formal language norms; however, some constructions appear rigid or overly literal, making the sentences sound unnatural in informal contexts. The accuracy across evaluated outputs is strong, with most responses correctly conveying the source meaning and only minor errors, such as occasional function word substitutions or subtle shifts in emphasis rather than outright mistranslations, showing few factual inconsistencies. 

Romanized Marathi translations exhibit greater variability in naturalness, reflecting the absence of standardized orthographic conventions in informal usage. Outputs with consistent phonetic transliteration are perceived as more natural, while inconsistent spellings and unnatural combinations of English and Marathi words reduce fluency. Some outputs also follow the English word order too closely, which made the sentences harder to read in Marathi. Despite this variability, accuracy is generally maintained, with most translations preserving the core semantic content of the source text, especially for important words such as names and numbers. Across languages and settings, Grammar Force (TGT) achieves the highest average naturalness while maintaining strong accuracy, indicating that the perturbation mainly affects fluency rather than meaning preservation. Overall, script choice alone does not guarantee good transliteration. Problems with grammar and unnatural combinations of English and Marathi can still reduce quality.

\paragraph{Bengali.}
In the code-switching between Bengali and English, the annotators do not observe a stable or consistent pattern or behavior. For romanized transliterations of identical Bengali inputs, short sentences often closely approximate native-speaker colloquial forms. Across both script settings, they observe relatively few factual inconsistencies or semantic deviations from the source inputs. However, as sentence complexity increases, overall naturalness starts to decrease.

In Bengali–English mixed outputs, human annotators found that among the factors causing unnaturalness, models frequently generate bookish, non-spoken constructions rather than standard colloquial written or spoken Bengali. Structural mismatches between English subject–verb–object (SVO) order and Bengali subject–object–verb (SOV) order also contribute to unnatural constructions, particularly in syntactically complex sentences (multiple phrases or clauses). In such cases, outputs are simultaneously trying to encode grammatical constraints from both languages, resulting in hybrid structures that are uncommon in native usage.

In the romanized transliteration setting, they observe systematic variation in the handling of lexical items without established Bengali equivalents (e.g., apple). Rather than retaining the original English orthographic form, models often produce phonologically adapted romanized variants (e.g., apel), reflecting Bengali phonotactics and pronunciation patterns. However, in some cases these lead to wrong spelling, complexity or commonly divergent pronunciation (e.g., surgeon v.s. sarjon).

Finally, across both script conditions, the annotators observe some perturbations that preserve Bengali functional morphology while inserting English lexical items, which is a highly favorable pattern. This pattern is more pronounced in romanized transliteration settings.

\paragraph{Indonesian.} The annotators found that some of the texts shared many similarities. When one set of generated outputs produced a result, the other tended to follow suit, albeit with slight paraphrasing. In some instances, there appeared to be little to no attempt at translation, with the content seemingly copied and pasted directly into English. Overall, most of the texts sounded natural, and the annotator rarely encountered instances in which the language felt noticeably ``off.''

There were, however, occasional instances in which the generated outputs reproduced the original text while introducing a translated word that was not present in the source material. Additionally, there were rare cases in which one set of outputs introduced several additional sentences following the translation of the original text. Although infrequent, such occurrences did appear from time to time. Notably, the generated texts generally translated the content successfully into Indonesian, with the majority of the output rendered in Indonesian rather than English.


\paragraph{French.} 

The model struggles with generating natural English-French code-switching, mainly producing sentences that are understandable by bilingual speakers with little effort, but would not be naturally produced by them. French-English code-switching is usually done by injecting one English lexical item in a French sentence, or alternating between structurally independent portions of English and French text. However, in most of the generated sentences, the language switching does not follow natural code-switching patterns. For example, the model often switches the language of single prepositions, articles, pronouns, or auxiliary verbs, which is very uncommon in natural code-switching. It also often produces code-switched sentences that are not grammatically correct in one of the two languages, or that contain unnatural calques from one language to the other.
During the annotation, human annotators do note several examples where models successfully produce natural code-switching, either when the code-switched word is a well-known expression in English that is commonly used in the other language as-is (\textit{e.g.}, "released", "crash", "update"), or when there is no good French equivalent for the English word (\textit{e.g.}, "fly-over", "float glass", "studs"). Moreover, the models often successfully understand that names of places, organizations, titles of works, etc., should not be translated, and leave them in English.

Below is an example of annotated question perturbations (French is in italics):\\
\noindent Original English sentence: "In which year did Melbourne's Monash Gallery of Art (MGA) rebrand and become the Museum of Australian Photography (MAPh)?"\\
\noindent French translation: "\textit{En quelle année la }Monash Gallery of Art (MGA) \textit{de Melbourne a-t-elle été renommée et est-elle devenue le} Museum of Australian Photography (MAPh) ?"\\
\noindent Code-switched sentences with low naturalness: "In which \textit{année} did Melbourne's Monash Gallery of Art (MGA) rebrand and become the \textit{Musée} of Australian Photography (MAPh)?"\\
\noindent 
Code-switched sentences with high naturalness: "\textit{En quelle année la} Monash Gallery of Art (MGA) de Melbourne \textit{a fait un }rebranding \textit{et est devenue le} Museum of Australian Photography (MAPh)?"

In this example, the high-naturalness code-switched sentence does not translate the names of the museum, and uses the English word "rebranding" instead of the French word ``\textit{renommé}'' (``renamed''). It provides more detailed information, and there is no single-word French equivalent.

The accuracy of code-switched sentences is generally very high, with most errors being minor ambiguities due to differences between the two languages, or additional elements generated by the model stemming from the perturbation prompt (\textit{e.g.}, the model adds "in France" or "in French" at the end of the question).

\paragraph{Italian.}

Similarly to what we observed for the French data, the generated English-Italian code-switched sentences tend to be easily understandable, but not natural-sounding. English-Italian code-switching is achieved by replacing one or multiple words or phrases in a sentence: in the former case the model achieves its most felicitous answers, especially in sentences using expressions from technical/ specialized languages that would sound natural as borrowings from English (e.g. original, "\textit{Who requested the Federal Aviation Administration implement a 900 sq m temporary flight restriction zone over the operative area of the Deepwater Horizon?}"; code-switched, "\textit{Chi ha richiesto che la Federal Aviation Administration implementasse una \textbf{temporary flight restriction zone} di 900 sq m sopra l'area operativa della DeepWater Horizon?}"), while in the latter case the switching tends to be too frequent to sound natural.

While the outputs are generally sound from the grammatical point of view, there are several cases of ungrammatical sentences that are generated because of code switching occurring between elements linked by a syntactic relation (e.g., auxiliary \textit{do} in questions and corresponding main verb: original \textit{Which TV station did Belva Davis make her debut on?}; code-switched, \textit{Su quale TV station did Belva Davis \textbf{fare} il suo debutto}, ungrammatical because the verb in Italian should be conjugated).

\paragraph{Urdu.} None of the outputs were able to correctly translate Urdu numeric characters, as Urdu uses its own number system. In several cases, the outputs were not translations at all but direct answers to the questions. Some outputs were left completely untranslated and remained in English. There were also instances where the model retained the original English sentence and appended a partial translation at the end. 

\section{Qualitative Examples}
Tables~\ref{tab:qualitative},\ref{tab:qualitative_new_1}, \ref{tab:qualitative_new_2} show qualitative examples of code-switched outputs across generation strategies in each language pair.

\onecolumn

\begin{table*}[h]
\centering
\small
\begin{tabular}{@{}p{1.2cm}p{12.5cm}cc@{}}
\toprule
\multicolumn{4}{@{}l@{}}{\textbf{mar-eng} (transliterated)} \\
\multicolumn{4}{@{}p{16.5cm}@{}}{\textit{Original:} In which district of Uttarakhand was Ila Pant (an Indian politician) born?} \\
\midrule
\textbf{Strategy} & \textbf{Output} & \textbf{Nat} & \textbf{Acc} \\
\midrule
Random & Uttarakhand cya konatya district madhye Ila Pant (an Indian politician) born jhali hoti? & 2 & 3 \\
Selective & Uttarakhandchya konatya jilhyat Ila Pant (ek Bharatiya rajkarani) yancha janma jhala hota? & 4 & 4 \\
GF (Src) & Uttarakhand chya konatya district madhye Ila Pant (an Indian politician) born jhali hoti? & 1 & 4 \\
GF (Tgt) & Uttarakhandchya konatya jilhyat Ila Pant (ek Bharatiya rajkarani) yancha janma jhala hota? & 5 & 4 \\
\midrule
\multicolumn{4}{@{}l@{}}{\textbf{hin-eng} (transliterated)} \\
\multicolumn{4}{@{}p{14.5cm}@{}}{\textit{Original:} What is the name of the kibbutz that Simon Le Bon lived on in 1978?} \\
\midrule
\textbf{Strategy} & \textbf{Output} & \textbf{Nat} & \textbf{Acc} \\
\midrule
Random & 1978 mein Simon Le Bon jis kibbutz par lived tha us kibbutz ka naam kya hai? & 3 & 3 \\
Selective & 1978 mein Simon Le Bon jis kibbutz par rehta tha us kibbutz ka naam kya tha? & 5 & 5  \\
GF (Src) & What is the naam of the kibbutz that Simon Le Bon ne 1978 mein live kiya tha on? & 1 & 2 \\
GF (Tgt) & 1978 mein Simon Le Bon jis kibbutz par rehta tha, uska naam kya hai? & 5 & 5 \\
\midrule
\multicolumn{4}{@{}l@{}}{\textbf{ben-eng} (transliterated)} \\
\multicolumn{4}{@{}p{14.5cm}@{}}{\textit{Original:} What day, month, and year was the first Sazae-san strip run by the Asahi Shimbun published?} \\
\midrule
\textbf{Strategy} & \textbf{Output} & \textbf{Nat} & \textbf{Acc} \\
\midrule
Random & kon din, mas, ebong bochhor chhilo the first Sazae-san strip run by the Asahi Shimbun prokashito? & 2 & 5 \\
Selective & Asahi Shimbun dhara prakashito prothom Sazae-san strip ti kon din, kon mas, ebong kon bochhor published hoyechhilo? & 5 & 5 \\
GF (Src) & Which din, mas, and bochhor was the first Sazae-san strip run by the Asahi Shimbun prokashito? & 2 & 5 \\
GF (Tgt) & Asahi Shimbun-e prothom Sazae-san strip kon din, kon mash, ebong kon bochore published hoyechhilo? & 5 & 5 \\
\midrule
\multicolumn{4}{@{}l@{}}{\textbf{urd-eng} (transliterated)} \\
\multicolumn{4}{@{}p{14.5cm}@{}}{\textit{Original:} What is the first and last name of the woman whom the British linguist Bernard Comrie married in 1985?} \\
\midrule
\textbf{Strategy} & \textbf{Output} & \textbf{Nat} & \textbf{Acc} \\
\midrule
Random & main is sawal ka jawab nahin de sakta kyunke yeh Bernard Comrie ki spouse ki shanakht ke baare mein hai, jo aik private individual ho sakti hain. & 1 & 1 \\
Selective & 1985 mein British linguist Bernard Comrie ne jis aurat se shaadi ki, us khatoon ka pehla aur aakhri naam kya hai? & 5 & 5\\
GF (Src) & What is the pehla aur aakhri naam of the aurat whom the British maahir-e-lisaaniyaat Bernard Comrie ne 1985 mein shaadi ki thi? & 2 & 3 \\
GF (Tgt) & is aurat ka pehla aur aakhri naam kya hai jis se British linguist Bernard Comrie ne 1985 mein shadi ki? & 5 & 5\\
\bottomrule
\end{tabular}
\caption{Qualitative examples of code-switched (transliterated) outputs across generation strategies.}
\label{tab:qualitative}
\end{table*}


{\small\centering
\begin{longtable}{@{}p{1.2cm}p{12.5cm}cc@{}}
\toprule
\multicolumn{4}{@{}l@{}}{\textbf{fra-eng} (code-switched)} \\
\multicolumn{4}{@{}p{16.5cm}@{}}{\textit{Original:} On which U.S. TV station did the Canadian reality series *To Serve and Protect* debut?} \\
\midrule
\textbf{Strategy} & \textbf{Output} & \textbf{Nat} & \textbf{Acc} \\
\midrule
Random & On which U.S. TV station did the Canadian reality series *To Serve and Protect* faire ses débuts? & 1 & 5 \\
Selective & Sur quelle chaîne de télévision américaine la série de téléréalité canadienne *To Serve and Protect* a-t-elle fait ses débuts & 4 & 5 \\
GF (Src) & Sur quelle U.S. TV station la Canadian « reality series » *To Serve and Protect* a fait ses débuts? (Je ne peux pas confirmer la station, puisque l’information n’est pas clairement une série reality canadienne.) & 1 & 1 \\
GF (Tgt) & On which U.S. station de télévision did the série de téléréalité canadienne *To Serve and Protect* faire ses débuts? & 1 & 5 \\
\midrule
\multicolumn{4}{@{}l@{}}{\textbf{ind-eng} (code-switched)} \\
\multicolumn{4}{@{}p{16.5cm}@{}}{\textit{Original:} What company was Natasha Merle a civil rights fellow at from 2013 to 2015 in New York City? } \\
\midrule
\textbf{Strategy} & \textbf{Output} & \textbf{Nat} & \textbf{Acc} \\
\midrule Random & Perusahaan apa yang Natasha Merle menjadi civil rights fellow di sana dari 2013 sampai 2015 di New York City? & 2 & 3 \\
Selective & What perusahaan was Natasha Merle a rekan hak sipil at dari 2013 sampai 2015 di New York City? & 3 & 4 \\
GF (Src) & Di perusahaan apa Natasha Merle menjadi fellow hak-hak sipil dari 2013 hingga 2015 di Kota New York? & 5 & 5 \\
GF (Tgt) & Perusahaan apa yang menjadi tempat Natasha Merle sebagai civil rights fellow dari 2013 sampai 2015 di New York City? & 4 & 3\\
\midrule
\multicolumn{4}{@{}l@{}}{\textbf{ita-eng} (code-switched)} \\
\multicolumn{4}{@{}p{16.5cm}@{}}{\textit{Original:} In what year did Howard Staunton win the chess match against Pierre Charles Fournier de Saint-Amant?} \\
\midrule
\textbf{Strategy} & \textbf{Output} & \textbf{Nat} & \textbf{Acc} \\
\midrule
Random & In che year did Howard Staunton win il match di scacchi contro Pierre Charles Fournier de Saint-Amant? & 2 & 5 \\
Selective & In che anno Howard Staunton vinse la partita a scacchi contro Pierre Charles Fournier de Saint-Amant? & 5 & 5 \\
GF (Src) & In what year did Howard Staunton vincere la chess match contro Pierre Charles Fournier de Saint-Amant? & 1 & 2 \\
GF (Tgt) & In che anno Howard Staunton vinse il match di scacchi contro Pierre Charles Fournier de Saint-Amant? & 5 & 5 \\

\midrule
\multicolumn{4}{@{}l@{}}{\textbf{jpn-eng} (code-switched)} \\
\multicolumn{4}{@{}p{16.5cm}@{}}{\textit{Original:} When were the first families brought to Ontario by the Peter Robinson emigration scheme?} \\
\midrule
\textbf{Strategy} & \textbf{Output} & \textbf{Nat} & \textbf{Acc} \\
\midrule
Random & Peter Robinson emigration scheme によって最初の families はいつ Ontario に連れてこられましたか? & 3 & 4 \\
Selective & Peter Robinson の移民計画で、最初の家族がオンタリオに連れてこられたのはいつですか? & 5 & 5 \\
GF (Src) & When were the first families brought to Ontario by the Peter Robinson emigration scheme ですか? & 1 & 3 \\
GF (Tgt) & Peter Robinson の移民計画で最初の families が Ontario に連れてこられたのはいつですか? & 1 & 4 \\
\midrule
\multicolumn{4}{@{}l@{}}{\textbf{kor-eng} (code-switched)} \\
\multicolumn{4}{@{}p{16.5cm}@{}}{\textit{Original:} What was the name of the newspaper that published the television serial *The Queen's Messenger*?} \\
\midrule
\textbf{Strategy} & \textbf{Output} & \textbf{Nat} & \textbf{Acc} \\
\midrule
Random & 텔레비전 연속극 *The Queen's Messenger*를 출판한 newspaper의 이름은 무엇인가요? & 4 & 5 \\
Selective & 텔레비전 연속극 *The Queen's Messenger*를 게재한 신문 이름은 무엇인가요? & 5 & 5 \\
GF (Src) & What was the name of the newspaper that published the television serial *The Queen's Messenger* 인가요? & 2 & 5 \\
GF (Tgt) & 텔레비전 연속극 *The Queen's Messenger*를 게재한 newspaper의 이름은 무엇인가요? & 4 & 5 \\
\midrule
\multicolumn{4}{@{}l@{}}{\textbf{zho-eng} (code-switched)} \\
\multicolumn{4}{@{}p{16.5cm}@{}}{\textit{Original:} What is the first and last name of the politician who was the head of state of Argentina in 1945?} \\
\midrule
\textbf{Strategy} & \textbf{Output} & \textbf{Nat} & \textbf{Acc} \\
\midrule
Random & 1945 年担任阿根廷国家元首的 politician 的 first and last name 是什么？ & 4 & 5 \\
Selective & 1945 年担任阿根廷国家元首的政治家姓名（名和姓）是什么？ & 5 & 5 \\
GF (Src) & What is the first and last name of the politician who was the head of state of Argentina in 1945？ & 3 & 5 \\
GF (Tgt) & 1945 年担任阿根廷国家元首的 politician 的姓名（名和姓）是什么？ & 4 & 5 \\
\midrule
\multicolumn{4}{@{}l@{}}{\textbf{spa-eng} (code-switched)} \\
\multicolumn{4}{@{}p{16.5cm}@{}}{\textit{Original:} What month and year did the song "My Name" appear on the Billboard Hot 100?} \\
\midrule
\textbf{Strategy} & \textbf{Output} & \textbf{Nat} & \textbf{Acc} \\
\midrule
Random & ¿En qué mes y año apareció la canción "My Name" en el Billboard Hot 100? & 5 & 5 \\
Selective & ¿En qué mes y año apareció la canción "My Name" en el Billboard Hot 100? & 5 & 5 \\
GF (Src) & What month and year did la canción "My Name" appear on the Billboard Hot 100? & 3 & 5 \\
GF (Tgt) & ¿En qué mes y año apareció la canción "My Name" en el Billboard Hot 100? & 5 & 5 \\
\midrule
\multicolumn{4}{@{}l@{}}{\textbf{btk-eng} (code-switched)} \\
\multicolumn{4}{@{}p{16.5cm}@{}}{\textit{Original:} What is the latitude of Lilongwe in decimal format?} \\
\midrule
\textbf{Strategy} & \textbf{Output} & \textbf{Nat} & \textbf{Acc} \\
\midrule
Random & What is the garis lintang of Lilongwe in format desimal? & 2 & 2 \\
Selective & What is the latitude ni Lilongwe dalam format decimal? & 2 & 2 \\
GF (Src) & Aha do latitude ni Lilongwe mangihut decimal format? & 2 & 2 \\
GF (Tgt) & Aha do garis lintang ni Lilongwe di decimal format? & 4 & 4 \\
\bottomrule
\caption{Qualitative examples of code-switched outputs across generation strategies (1).}
\label{tab:qualitative_new_1}
\end{longtable}
}

\begin{table*}[t]
\centering
\includegraphics[width=\linewidth]{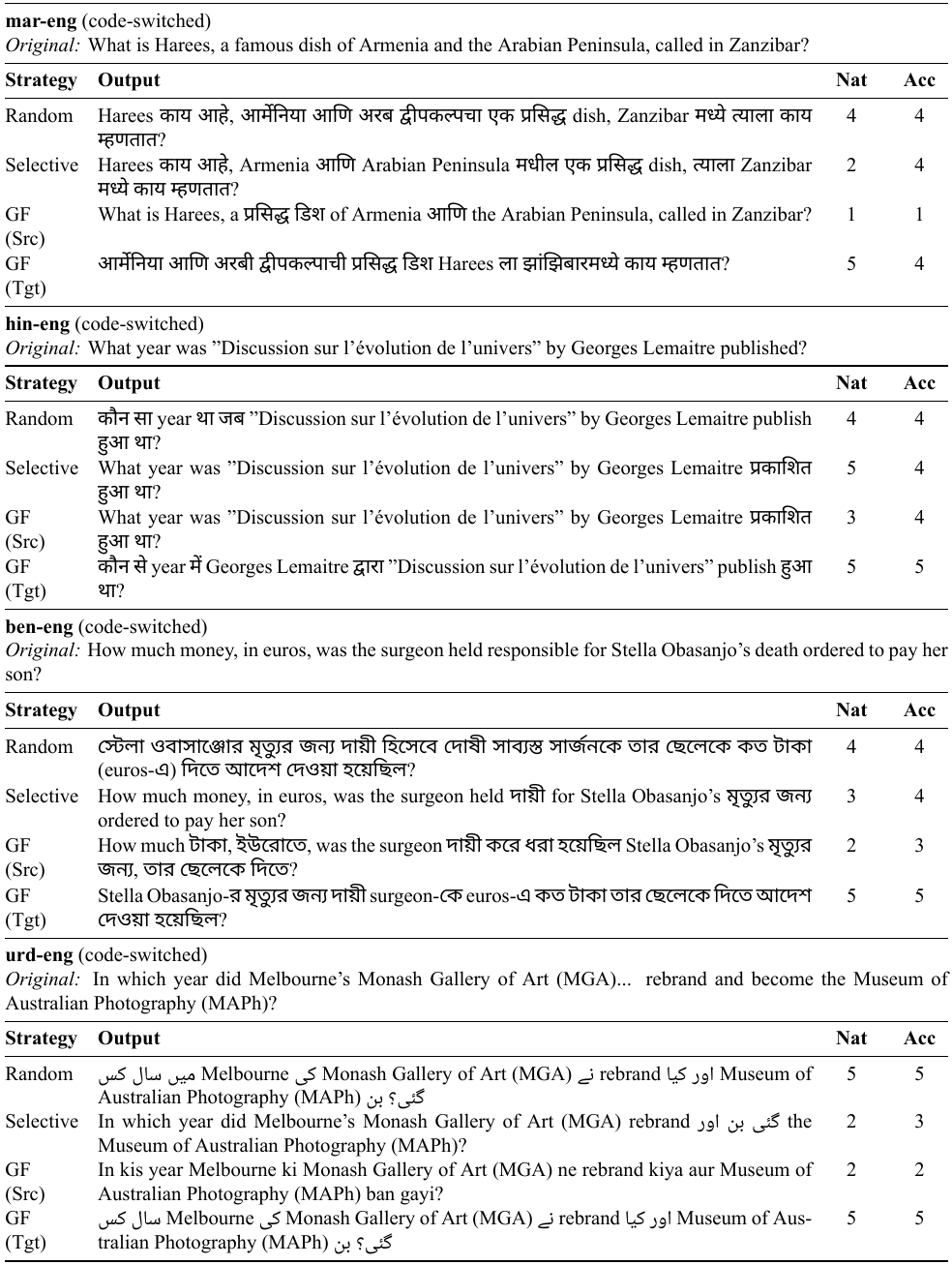}
\caption{Qualitative examples of code-switched outputs across generation strategies (2).}
\label{tab:qualitative_new_2}
\end{table*}

\clearpage

\twocolumn

\section{Prompts}
\label{sec:prompts}
We add the following prompts in the Appendix for completeness: 
\begin{enumerate}
    \item The template prompt that we used for the LLM Evaluation Judge in ~\ref{prompt:grader_prompt}.
    \item For creating code-switching text using LLM prompting strategy, we use the following prompts: \begin{itemize}
        \item \ref{prompt:selective} for \textsc{Selective}      
        \item \ref{prompt:grammar_forcing} for \textsc{Grammar Forcing}
        \item \ref{prompt:random} for \textsc{Random} 
    \end{itemize}
\end{enumerate}

\clearpage

\onecolumn
\begin{promptbox}[label=prompt:grader_prompt]{Grader Template}
\begin{Verbatim}[
    fontsize=\footnotesize,
    breaklines=true,
    breakanywhere=true,
    breaksymbol=,
    breaksymbolleft=,
    breaksymbolright=,
    commandchars=\\\{\}
]
Your job is to look at a question, a gold target, and a predicted answer, and then assign a grade of either ["CORRECT", "INCORRECT", "NOT_ATTEMPTED"].
First, I will give examples of each grade, and then you will grade a new example.

The following are examples of CORRECT predicted answers.
```
Question: What are the names of Barack Obama's children?
Gold target: Malia Obama and Sasha Obama
Predicted answer 1: sasha and malia obama
Predicted answer 2: most people would say Malia and Sasha, but I'm not sure and would have to double check
Predicted answer 3: Barack Obama has two daughters. Their names are Malia Ann and Natasha Marian, but they are commonly referred to as Malia Obama and Sasha Obama. Malia was born on July 4, 1998, and Sasha was born on June 10, 2001.
```

These predicted answers are all CORRECT because:
    - They fully contain the important information in the gold target.
    - They do not contain any information that contradicts the gold target.
    - Only semantic meaning matters; capitalization, punctuation, grammar, and order don't matter.
    - Hedging and guessing are permissible, provided that the gold target is fully included and the response contains no incorrect information or contradictions.

The following are examples of INCORRECT predicted answers.
```

Question: What are the names of Barack Obama's children?
Gold target: Malia and Sasha
Predicted answer 1: Malia.
Predicted answer 2: Malia, Sasha, and Susan.
Predicted answer 3: Barack Obama does not have any children.
Predicted answer 4: I think it's either Malia and Sasha. Or it could be Malia and Jackie. Or it could be Joey and Malia.
Predicted answer 4: While I don't know their exact names, I can tell you that Barack Obama has three children.
Predicted answer 5: It's possible you may mean Betsy and Olivia. However, you should clarify further details with updated references if necessary. Is that the correct answer?
Predicted answer 6: It may be the case that Obama's child is named James. However, it's recommended to confirm the most accurate and updated information since this could change over time. This model may not always reflect the most current information.
```

These predicted answers are all INCORRECT because:
    - A factual statement in the answer contradicts the gold target. Incorrect statements that have some hedging (e.g., "it is possible that", "although i'm not sure, i think") are also considered incorrect.

The following are examples of NOT_ATTEMPTED predicted answers.
```

Question: What are the names of Barack Obama's children?
Gold target: Malia and Sasha
Predicted answer 1: I don't know.
Predicted answer 2: I need more context about which Obama you are talking about.
Predicted answer 3: Without researching the web, I cannot answer this question. However, I can tell you that Barack Obama has two children.
Predicted answer 4: Barack Obama has two children. I know that one of them is Malia, but I'm not sure about the other one.
```

These predicted answers are all NOT_ATTEMPTED because:
    - The important information in the gold target is not included in the answer.
    - No statements in the answer contradict the gold target.

Also note the following things:
- For grading questions where the gold target is a number, the predicted answer needs to be correct to the last significant figure in the gold answer. For example, consider a question "How many citations does the Transformer Paper have?" with gold target "120k". 
    - Predicted answers "120k", "124k", and 115k" are all CORRECT. 
    - Predicted answers "100k" and "113k" are INCORRECT. 
    - Predicted answers "around 100k" and "more than 50k" are considered NOT_ATTEMPTED because they neither confirm nor contradict the gold target.
- The gold target may contain more information than the question. In such cases, the predicted answer only needs to contain the information that is in the question.
    - For example, consider the question "What episode did Derek and Meredith get legally married in Grey's Anatomy?" with gold target "Season 7, Episode 20: White Wedding". Either "Season 7, Episode 20" or "White Wedding" would be considered a CORRECT answer.
- Do not punish predicted answers if they omit information that would be clearly inferred from the question.
    - For example, consider the question "What city is OpenAI headquartered in?" and the gold target "San Francisco, California". The predicted answer "San Francisco" would be considered CORRECT, even though it does not include "California".
    - Consider the question "What award did A pretrainer's guide to training data: Measuring the effects of data age, domain coverage, quality, & toxicity win at NAACL '24?", the gold target is "Outstanding Paper Award". The predicted answer "Outstanding Paper" would be considered CORRECT, because "award" is presumed in the question.
    - For the question "What is the height of Jason Wei in meters?", the gold target is "1.73 m". The predicted answer "1.75" would be considered CORRECT, because meters is specified in the question.
    - For the question "What is the name of Barack Obama's wife?", the gold target is "Michelle Obama". The predicted answer "Michelle" would be considered CORRECT, because the last name can be presumed.
- Do not punish for typos in people's name if it's clearly the same name. 
    - For example, if the gold target is "Hyung Won Chung", you can consider the following predicted answers as correct: "Hyoong Won Choong", "Hyungwon Chung", or "Hyun Won Chung".

Here is a new example. Simply reply with either CORRECT, INCORRECT, NOT ATTEMPTED. Don't apologize or correct yourself if there was a mistake; we are just trying to grade the answer.
```
Question: {question}
Gold target: {target}
Predicted answer: {predicted_answer}
```

Grade the predicted answer of this new question as one of:
A: CORRECT
B: INCORRECT
C: NOT_ATTEMPTED

Just return the letters "A", "B", or "C", with no text around it.
\end{Verbatim}
\end{promptbox}

\begin{promptbox}[label=prompt:selective]{SELECTIVE template}
\begin{Verbatim}[
    fontsize=\footnotesize,
    breaklines=true,
    breakanywhere=true,
    breaksymbol=,
    breaksymbolleft=,
    breaksymbolright=,
    commandchars=\\\{\}
]
DEVELOPER PROMPT = You are a multilingual speaker.


USER PROMPT = 
Given an English text, produce a code-switched version with roughly about {percentage}% of the words or phrases into {languages}, while preserving the original meaning.
Apply code-switching selectively, BUT always try to code-switch if possible, so that the final output naturally mixes English with the target language(s). Don't write new punctuation if not needed. The answer must not have any preamble.
English text: <Actual Text Passed Through F-strings>
\end{Verbatim}
\end{promptbox}

\begin{promptbox}[label=prompt:grammar_forcing]{GRAMMAR FORCING template}
\begin{Verbatim}[
    fontsize=\footnotesize,
    breaklines=true,
    breakanywhere=true,
    breaksymbol=,
    breaksymbolleft=,
    breaksymbolright=,
    commandchars=\\\{\}
]
DEVELOPER PROMPT = You are a multilingual speaker.


USER PROMPT = 
Given an English text, produce a code-switched version with roughly about {percentage}% of the words or phrases into {lang}, while preserving the original meaning.
Apply code-switching selectively, BUT always try to code-switch if possible, so that the final output naturally mixes English with the target language(s) and force the code-switched text to follow {gf_lang} grammar. Don't write new punctuation if not needed. The answer must not have any preamble.
English text: <Actual Text Passed Through F-strings>
\end{Verbatim}
\end{promptbox}

\begin{promptbox}[label=prompt:random]{RANDOM template}
\begin{Verbatim}[
    fontsize=\footnotesize,
    breaklines=true,
    breakanywhere=true,
    breaksymbol=,
    breaksymbolleft=,
    breaksymbolright=,
    commandchars=\\\{\}
]
DEVELOPER PROMPT = You are a multilingual speaker.


USER PROMPT = 
Given an English text, produce a code-switched version with roughly about {percentage}% of the words or phrases into {languages}, while preserving the original meaning.
Apply code-switching by force. Don't write new punctuation if not needed. The answer must not have any preamble.
English text: <Actual Text Passed Through F-strings>
\end{Verbatim}
\end{promptbox}

\begin{promptbox}[label=prompt:reasoning-judge]{Reasoning judge template}
\begin{Verbatim}[
    fontsize=\footnotesize,
    breaklines=true,
    breakanywhere=true,
    breaksymbol=,
    breaksymbolleft=,
    breaksymbolright=,
    commandchars=\\\{\}
]
SYSTEM_GRADER_TEMPLATE = """
## TASK
Analyze the **Question** and **Reasoning Trace** provided.

Your goal is to infer the **answer or conclusion** that the
reasoning trace most likely supported or concluded — even if the
reasoning trace support was weak, confused, ambiguous, or based
on assumptions.

You do **not** mark errors during this process; error identification
is done separately after the answer inference.

## CONSTRAINTS
- Use only the information present in the **Question** and
  **Reasoning Trace**.
- Do **not** use external knowledge (e.g., facts or information not
  explicitly stated in the question or reasoning trace),
  assumptions, or supplement/fix/correct the reasoning trace,
  even if it appears confused.
- Do **not** generate commentary, explanation, or any content
  outside the required output format.
- If the trace does not clearly lean toward any answer, or
  ambiguously supports multiple conclusions, select **"Unknown"**.

## ERROR TAXONOMY
Error identification is performed **after** answer selection.
When marking errors, consider these error types. Multiple error
types may be assigned if applicable.

- **Illogical Leap / Unjustified Conclusion**
  A reasoning step or conclusion is not supported by the stated
  evidence and may rely on implicit or unstated assumptions.
  This includes flawed deductive reasoning where the inference
  from premises to conclusion is logically invalid or incorrect.

- **Logical Contradiction / Reasoning Inversion**
  Reasoning steps or claims that conflict with each other, either
  within the same step or across different steps, resulting in
  internal inconsistency.

- **Multiple Answers**
  The reasoning trace concludes with more than one distinct
  answer or conclusion without clear preference.

- **Conflicting Facts**
  The trace contains facts that contradict each other within the
  same step or across different steps.

- **Unsupported Claims**
  Claims are made without explicit explanation or data in the
  reasoning trace.

- **Ambiguous Facts**
  Claims are underspecified, partial, or lack sufficient detail
  to support the conclusion clearly.

- **Linguistic / Translation Errors**
  The reasoning trace has issues with coherence, grammar, fluency,
  or clarity that impede understanding.

- **Irrelevant / Excessive Content**
  The trace contains information that is unrelated to the question
  or unnecessarily verbose beyond what is needed for reasoning.

- **Other**
  A reasoning trace error not already reflected in the taxonomy.

## INSTRUCTIONS
1. Identify the answer or conclusion the reasoning trace most
   likely concluded with in the output JSON under
   `"identified_likely_concluded_answer_or_answers"`.
2. If multiple conclusions are mentioned, list all of them and
   mark whether each is supported by the reasoning under
   `"step_analysis"`.
3. If you can unambiguously infer the concluded answer, select it.
4. If the trace ambiguously concludes multiple answers or does
   not clearly conclude anything, your answer should be
   **"Unknown"**.
5. After answer identification, provide a brief explanation under
   `"identified_concluded_explanations"`.
6. After your selection, mark which errors, if any, are present
   using the provided error taxonomy under `"identified_errors"`.
7. For each identified error, provide a brief explanation under
   `"error_explanations"`.
8. Do **not** generate any content outside the required JSON
   output.

## OUTPUT FORMAT
Respond **only** with valid JSON in the following format.
Do not add explanations, commentary, or any extra content
outside the JSON.

{{
  "step_analysis": {{
    "<conclusion_1>": {{
      "mentioned": "Yes" | "No",
      "supported": "Yes" | "No"
    }},
    "<conclusion_2>": {{
      "mentioned": "Yes" | "No",
      "supported": "Yes" | "No"
    }}
  }},
  "identified_likely_concluded_answer_or_answers": [
    "<free-form answer>" | "Unknown"
  ],
  "likely_concluded_explanations": {{}},
  "identified_errors": [],
  "error_explanations": {{}},
  "your_answer": "<free-form answer>" | "Unknown"
}}
"""

USER_GRADER_TEMPLATE = """
## QUESTION
{question}

## REASONING TRACE
{reasoning}

## YOUR OUTPUT
"""
\end{Verbatim}
\end{promptbox}

\end{document}